\documentclass{article}

\usepackage{microtype}
\usepackage{graphicx}
\usepackage{booktabs} 

\usepackage{hyperref}

\usepackage[final]{ARLET_2024}

\usepackage{amsmath}
\usepackage{amssymb}
\usepackage{mathtools}
\usepackage{amsthm}

\usepackage{array} 
\newcolumntype{C}[1]{>{\centering\arraybackslash}p{#1}}

\usepackage[capitalize,noabbrev]{cleveref}

\theoremstyle{plain}

\theoremstyle{definition}

\theoremstyle{remark}

\usepackage[textsize=tiny]{todonotes}

\usepackage{xurl} 

\usepackage{graphicx}
\usepackage{enumitem}
\usepackage{listings}
\usepackage{xcolor}
\usepackage{authblk}
\usepackage{lipsum}
\usepackage{threeparttable} 
\usepackage{subcaption} 
\definecolor{completioncolor}{rgb}{0.5,0,0} 

\title{Jack of All Trades, Master of Some, a Multi-Purpose Transformer Agent}
\author[1,2]{Quentin Gallouédec}
\author[1]{Edward Beeching}
\author[1,3]{Clément Romac}
\author[2]{Emmanuel Dellandréa}

\affil[1]{Hugging Face}
\affil[2]{Ecole Centrale de Lyon, CNRS, Universite Claude Bernard Lyon 1, INSA Lyon, Université Lumière Lyon 2, LIRIS, UMR5205, 69130 Ecully, France}
\affil[3]{Inria (Flowers), University of Bordeaux, France}

\begin{document}

\maketitle

\begin{abstract}
The search for a general model that can operate seamlessly across multiple domains remains a key goal in machine learning research. 
The prevailing methodology in Reinforcement Learning (RL) typically limits models to a single task within a unimodal framework, a limitation that contrasts with the broader vision of a versatile, multi-domain model.
In this paper, we present Jack of All Trades (JAT), a transformer-based model with a unique design optimized for handling sequential decision-making tasks and multi-modal data types.
The JAT model demonstrates its robust capabilities and versatility by achieving strong performance on very different RL benchmarks, along with promising results on Computer Vision (CV) and Natural Language Processing (NLP) tasks, all using a single set of weights.
The JAT model marks a significant step towards more general, cross-domain AI model design, and notably, it is the first model of its kind to be fully open-sourced\footnote{\url{https://huggingface.co/jat-project/jat}},
including a pioneering general-purpose dataset.

\end{abstract}

\section{Introduction}

Machine learning researchers have long aimed to develop versatile models that can adapt seamlessly to different domains.
The recent success of Transformers \citep{vaswani2017attention} in NLP, CV, and to some extent in RL, has opened new avenues in this quest. 
In this paper, we attempt to extend the boundaries of this success by proposing a single, unified model capable of operating across a wide range of NLP, CV, and RL tasks using a single set of parameters. This effort not only seeks to challenge the conventional compartmentalization of AI tasks into distinct domains, but also aims to establish a more holistic approach to AI model design.

While combining visual and textual tasks has been well-researched, integrating RL tasks remains relatively unexplored and poses distinct challenges.
RL tasks are inherently diverse and heterogeneous, making their combination among themselves and with other domains a highly complex exercise. This integration requires dealing with a landscape of different modalities, task complexities, and data volumes across domains and tasks. New questions that arise include: (1)~How to design a model and learning method that effectively handles different modalities and data types (sequential decision-making and text-centric)? (2)~How to formulate a learning objective that appropriately balances and harmonizes the different modalities, tasks, and domains without bias toward any particular domain or task? (3)~How to design a learning strategy that can accommodate the different levels of complexity inherent in different tasks? 

These goals are concurrent and, to our knowledge, have only been addressed together by \citet{reed2022generalist} with the Gato model. 
Our contributions are characterized by three major advances:
(1) Our model features an innovative structure optimized for sequential decision-making tasks. It uniquely assigns each timestep to a corresponding token embedding, resulting in a simpler design. 
This approach significantly expands the attention window in terms of timesteps compared to Gato (e.g., it is 19 times larger for Atari and more than 25 times larger for Meta-World).
(2) In the spirit of open source, we release our code, dataset, and model to the research community.
(3) We add observation prediction as an auxiliary task to our model. We demonstrate that this integration significantly contributes to learning a more efficient agent.

Ultimately, our JAT model achieves competitive results on the tasks studied, while being more than 6 times smaller than Gato and relying on a significantly lower training budget.
As mentioned above, this new paradigm raises a number of open questions and paves the way for new research. We present a first milestone in this emerging framework and acknowledge the significant potential for improved results.

\section{Related Work}
\subsection{Transformer for RL}
Transformer models \citep{vaswani2017attention} are designed to model sequences and in particular sequences of words in natural language. 
However, sequence modeling problems span over a much larger set of domains than only NLP. Several efforts have been made to leverage these models for RL \citep{li2023survey}. In this paper, we focus on modeling RL trajectories (i.e., sequences of  observations, actions and rewards). Modeling such sequences with a Transformer was introduced by \citet{chen2021decision} and the Decision Transformer (DT) model. In DT, a Transformer model is trained with offline RL to take sequences of transitions as input and predict the next action. In particular, \citet{chen2021decision} proposed to use returns-to-go (i.e. the return from the current state) to condition actions' generation on both the previous observation and the desired return-to-go. While this has the advantage of explicitly modeling the relations between action selection and return, using the model at inference requires providing at every step a desired return. \citet{liu2023emergent} proposed to extend this by using hindsight relabelling to better exploit sub-optimal trajectories. \citet{zheng2022online} also extended the DT approach by mixing offine pretraining and online finetuning. Finally, \citet{lee2022multi} studied how the DT approach scales to a multi-task RL setup where a single policy is learned for multiple games.
In terms of sequence structure, some discretize each dimension of the observation and action spaces separately \citep{reed2022generalist, janner2021offline, chebotar2023q}, while others associate an embedding with each element of the sequence \citep{chen2021decision, zheng2022online}.

Our work lies in this line of work as it also leverages Transformers to model trajectories. However, our approach 
(1) uses standard Behavior Cloning (BC) instead of conditional BC, relaxing the need to condition the agent by the return-to-go and 
(2) models a multi-task dataset in which sequences come from very different domains (e.g. control, Atari, visual question answering, see Section \ref{sec:dataset}). 

\subsection{Multi-Modal Transformer}

Apart from being widely used in NLP, Transformers also thrive in vision and vision-and-language domains. As one of the first works leveraging Transformers for vision, \citet{dosovitskiy2021image} introduced Vision Transformer (ViT), a Transformer model using image patches for recognition. Following this, a line of work aiming to train multi-modal Transformers using both text and images emerged, including works such as Flamingo \citep{alayrac2022flamingo}, PaLI \citep{chen2023pali} or IDEFICS \citep{laurençon2023obelisc}. All these models imply the use of an image encoder allowing to obtain image tokens or embeddings that can be given to the Transformer alongside text tokens.


These models, typically generating text outputs, are trained for vision-and-language tasks like Visual Q\&A. However, recent multi-modal Transformers focus on decision-making. For example, \citet{jiang2022vima} trained a robot with Imitation Learning (IL) using multi-modal prompts to produce motor actions. RT-1 \citep{brohan2023rt1} and RT-2 \citep{brohan2023rt2} use expert demonstrations for real-world robots, with RT-2 building on RT-1 by directly outputting motor actions. Palm-E \citep{driess2023palm} leverages a pretrained Visual Language Model (VLM) for robotics tasks, producing sequences of text instructions executed by control policies.

Finally, our approach is largely inspired by Gato \citep{reed2022generalist}, which proposed to train a Transformer on both vision-and-language and decision-making tasks without relying on any pretrained model. The resulting model is therefore smaller than the ones leveraging large VLMs (e.g. Palm-E, RT-2) while still being able to perform both vision-and-language and decision-making tasks. In this paper, we first propose to build a dataset that resembles Gato's dataset except we only use open-source data sources and release all demonstrations as well as expert policies we used to obtain these demonstrations. Then, we also leverage a multi-modal Transformer along with IL for our model, but introduce several improvements, notably in the processing of sequential data and support for continuous values
(see Section \ref{subsec:architecture}).

\subsection{Multi-Task RL}

The quest for a general agent has long been a goal of RL \citep{bellemare2013arcade}. However, most works have chosen to use a different neural network for each environment. Recent research has revived interest in this objective and explores it through several approaches.

One such approach involves directly extending online learning to multi-task environments \citep{espeholt2018impala, yu2019meta, song2020v}. These works highlight the potential for positive transfer in multi-task learning, meaning that learning across tasks can be mutually beneficial due to underlying commonalities. However, they also acknowledge the risk of negative transfer, where inter-task interference can impair training. 
Studies have investigated methods to limit this risk, such as that by \citet{yang2020multi}, which proposed refined gradient management techniques to mitigate these detrimental effects.

An alternative approach is policy distillation, which involves condensing the behaviors of expert agents into a singular, unified policy \citep{rusu2016policy, parisotto2016actor}.
While these studies also report positive transfer across tasks \citep{rusu2016policy}, they also identify instances of negative transfer. Subsequent research has focused on strategies to minimize this negative transfer \citep{teh2017distral}. 
The reliance on the availability of policies to distill is a limitation.
This constraint is notably addressed in \citep{chen2021decision}, which proposes conditioning the distilled policy on the desired return thus allowing the use of any policy, including those of non-expert agents. This strategy has been adapted to the multi-task setting by \citet{lee2022multi}.

Despite the diversity of research in this area, most studies are limited to multi-task learning within a single domain, such as Atari or Meta-World, and thus involve semantically related tasks (although this is somewhat less true for Atari). 
The only notable exception we found is the Gato model \citep{reed2022generalist}, which learns a large number of domains in a single network. It is the closest baseline to our work. 

\section{Methodology}

In this section, we introduce the JAT model, detailing our architectural choices that underpin its effectiveness and highlighting its ability to handle different modalities in both sequential decision-making and text-centric tasks. We present the associated dataset, which is notable for its groundbreaking diversity across domains and modalities. Finally, we discuss in depth the learning strategy used.

\subsection{Model Architecture} \label{subsec:architecture}

\begin{figure}[ht]
\vskip 0.2in
\begin{center}
\centerline{\includegraphics[width=0.6\textwidth]{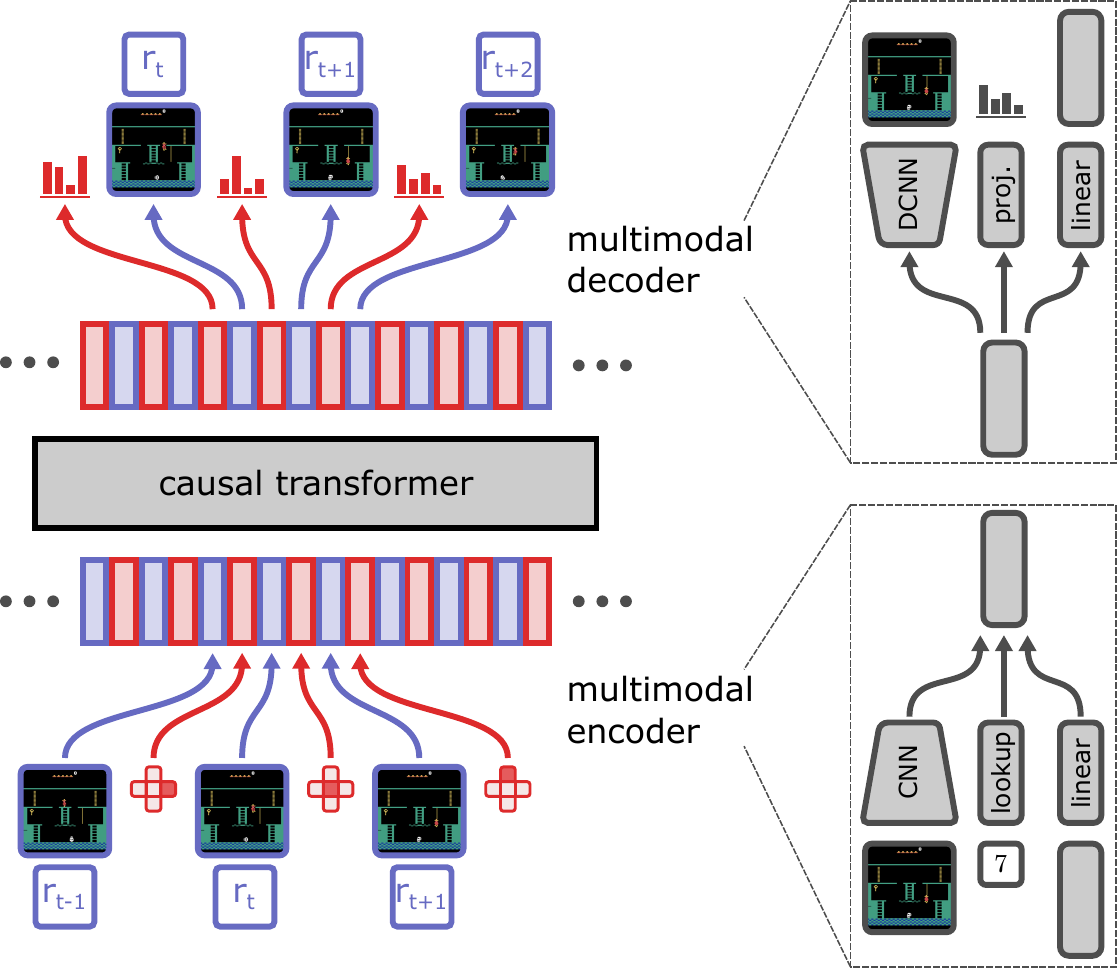}}
\caption{Architecture of the JAT network. For sequential decision-making tasks, observations and rewards on the one hand, and actions on the other, are encoded and interleaved. The model generates the next embedding autoregressively with a causal mask, and decodes according to expected modality.}
\label{fig:model}
\end{center}
\vskip -0.2in
\end{figure}

\subsubsection{Embedding Mechanism}

The model is designed to handle two main categories of data: tasks involving sequential decision-making and text-centric tasks. 
In text-centric tasks, the model currently supports two modalities: text and image. Although the current version of the model supports image generation, we focus only on tasks that involve text generation. To ease reading, we will refer to it as \textit{text-centric tasks} in the remainder of this paper.
Each of these two categories requires a slightly different approach to the embedding process.

In both cases, the resulting sequence is truncated to match the maximum permissible input size of the inner Transformer model. Any truncated portion is not discarded; instead, it forms the basis of a new sample. This process may be repeated if necessary, ensuring that no valuable information is lost.

\paragraph{Sequential Decision-Making Tasks}
For sequential decision-making tasks, the data comprises a sequence of observations, actions, and rewards. At the embedding stage, these sequences are processed to produce an interleaved sequence of observation embeddings (augmented with the corresponding reward) and action embeddings, denoted as $[\phi(s_0, 0.0), \phi(a_0), \phi(s_1, r_1), \phi(a_1), \ldots]$. Unlike DT \citep{chen2021decision} and Gato \citep{reed2022generalist}, each timestep is consistently associated with two embeddings: one for the observation and the other for the action, regardless of the modality. 
This enables JAT to better handle high-dimensional observations, and to provide a much wider, constant attention window in terms of timesteps.
As an example, this multiplies the size of the attention window in terms of timesteps by more than 25 for Meta-World.
The embedding method employed at a specific timestep is modality-dependent (with $H$ the hidden size of the model):

\begin{itemize}[noitemsep,topsep=0pt,left=1pt]
    \item Continuous observation: The reward value is appended to the observation vector. This augmented vector is then padded to achieve a uniform length of 377, corresponding to the maximum augmented observation size in the dataset. The embedding vector is subsequently obtained by passing this padded vector through a linear layer with an output size of $H$. This layer is consistently used across all timesteps.
    \item Discrete observation: The observation consists of a vector of integers, each of which is encoded into a continuous vector of size $H$ using a lookup table. Subsequently, a linear layer is applied to reduce the dimensionality to $\lfloor H/50\rfloor$. Following vector flattening, another linear layer is applied, resulting in an output size of $H-1$. Lastly, the reward is added to the resulting vector.
    \item Image observation: The input image is first resized to a uniform dimension of $84\times 84$ using bicubic approximation, normalized, and padded to ensure 4 channels. The image encoder consists of a series of three blocks, each consisting of a convolutional layer, an instance normalization layer, and an attention layer. The output of the last block is flattened and passed through a linear layer, resulting in an embedding vector of size $H$.
    \item Continuous action: The process is similar to that of continuous observations, with the exception of the reward component. Notably, the linear layer is shared with the one used for continuous observations.
    \item Discrete action: In the case of discrete actions, the process is slightly different due to the nature of the input: a discrete action is represented by a single integer, as opposed to a vector of integers for discrete observations. The input is directly mapped to a continuous vector of size $H$ using the same lookup table employed for discrete observations.
\end{itemize}

\paragraph{Text-Centric Tasks}
For text-centric tasks, each sample includes text, accompanied or not by an image.

\begin{itemize}[noitemsep,topsep=0pt,left=1pt]
    \item Image data: We employ the ViT architecture, as originally proposed by \citet{dosovitskiy2021image}. The image is first cropped to its central square, and resized to $224 \times 224$. The image is then normalized and divided into non-overlapping patches of $16 \times 16$. Each patch is linearly embedded in a vector of size $H$.
    \item Text data: We use the GPT-2 tokenization strategy \citep{radford2019language}, utilizing a byte-pair encoding \citep[BPE]{sennrich2016neural} specifically designed for unicode characters. This approach ensures comprehensive and granular tokenization. The tokenizer produces a vocabulary of 50,257 tokens. For efficient implementation, we use the Hugging Face integration \citep{moi2023huggingface}. Each token is mapped to an embedding vector using a lookup table, where each unique token in the vocabulary is associated with a distinct vector. Notably, this lookup table is shared with the one employed for discrete values in sequential decision-making tasks. 
\end{itemize}

When a sample includes both images and text, the embeddings are arranged so that the image embeddings precede the text embeddings. This specific order is essential for image captioning task because of the causal masking applied by the model's internal Transformer. The concatenated image-text embeddings form a unified representation for subsequent processing steps.

\subsubsection{Transformer Architecture}

The JAT model is based on a Transformer architecture using EleutherAI's implementation of GPT-Neo \citep{black2021gpt}.
It takes as input the embedding sequence whose computation was described in the previous section. The model uses a dual attention mechanism whose design is inspired by the Longformer \citep{beltagy2020longformer}: global attention with a window size of 512 tokens for full context understanding, and local attention with a fixed window of 256 tokens. 
The Transformer's feed-forward components consist of 12 layers and 12 heads with an intermediate dimensionality of 8192 and a hidden size of 768. They are designed to be causal, meaning a causal mask is applied during training and inference.

\subsubsection{Output Processing and Loss}

The internal causal Transformer outputs a sequence of embeddings, each encoding the basis for predicting subsequent elements in different data modalities. As we predict multiple modalities within a single sequence, we use the appropriate decoders and corresponding loss functions for each modality.
When an embedding encodes an image, we use a transposed convolutional neural network \citep{zeiler2010deconvolutional} for prediction. When an embedding represents a continuous vector, we use a continuous linear layer for prediction. For both image and continuous vector prediction, we compute the loss using Mean Square Error (MSE).
When an embedding represents a discrete value, we assign scores to each discrete candidate using a linear projection layer and compute the loss using cross-entropy.
Notably, we use the same projection layer for text tokens and discrete sequential values (like action for Atari and BabyAI).
To compute the overall loss of the sequence, we average the individual losses computed for each element. For sequential decision-making task, we apply a weighting between the loss related to observations and the loss related to actions. We show in Section \ref{subsec:ablation_obs_loss} that predicting the observations does help learning, and thereby solve one of the common open questions of \citep{chen2021decision} and \citep{reed2022generalist}.

\subsection{Datasets}
\label{sec:dataset}

In this work, we have collected a wide range of datasets, classified into two main groups: sequential decision-making datasets and textual datasets. The former include a series of interaction sequences, each consisting of observations, actions and a subsequent rewards, generated by so-called expert agents, details of which are given in the Appendix \ref{appendix:dataset}. The latter includes large corpora of textual data and image-text pairs.
In order to promote the emerging field of general-purpose AI models, we have made these datasets, together with the expert agents and the full set of code required to generate them, available to the public as open resources in our Hugging Face repository, accessible at the following URL \url{https://huggingface.co/jat-project}.
To the best of our knowledge, this compilation is unprecedented in terms of the variety of tasks and the volume of data, representing a valuable new contribution to the field.

\subsection{Training}

\subsubsection{Overall Training Procedure}

The model was trained for 250,000 steps. We distributed the training across 8 GPUs NVIDIA V100 using the Trainer from the Hugging Face Transformers library \citep{wolf2020transformers} in conjunction with Accelerate \citep{gugger2022accelerate}. This training lasted approximately 9 days.
For practical reasons, each batch is made up of data from a single dataset.
We use a constant batch size of 20 and accumulate over 2 steps, resulting in an effective batch size of 320. 
We use the AdamW optimizer with parameters $\beta_1=0.9$, $\beta_2=0.999$, and $\epsilon=10^{-8}$. The learning rate starts at $5\cdot 10^{-5}$ and linearly decays to zero throughout the training process.

\subsubsection{Task-Specific Weight Adjustments}
Each task presents a unique training challenge. To allow for balanced learning of all tasks, we introduced custom weight modifications. The choice of these weights is made heuristically. The learning would surely benefit from a more precise and systematic method for choosing these weights.

\textbf{Sample Weight} Some tasks required more updates for effective convergence. To allow proportionate progress of all tasks during learning, these tasks are sampled more frequently. Specifically, Oscar, Conceptual-Captions and Wikipedia was assigned a sample weight of 10.0 while other have a sample weight of 1.0.

\textbf{Loss Weight} Some control tasks require increased accuracy of actions. To allow for more strongly penalizing the error for these tasks, we assigned loss weights.
In MuJoCo tasks, the loss weight is typically set at 10.0, except for the Pendulum task (20.0) and the Double Pendulum task (50.0). In Meta-World tasks, a uniform loss weight of 50.0 is used.

\section{Experiments and Results}

In this section we discuss the results of our experiments. First, we provide a brief overview of the model's performance on text-centric tasks. We then present the results of the sequential decision-making tasks, showing the different levels of mastery across the different domains within our study. Finally, we provide a comprehensive analysis highlighting the benefits of incorporating the prediction of the next observation as an auxiliary task during the learning process.

\subsection{Text-Centric Tasks}

\begin{figure}[ht]
\centering
\begin{tabular}{C{4.3cm}C{4.3cm}C{4.3cm}}
\includegraphics[width=0.17\textwidth]{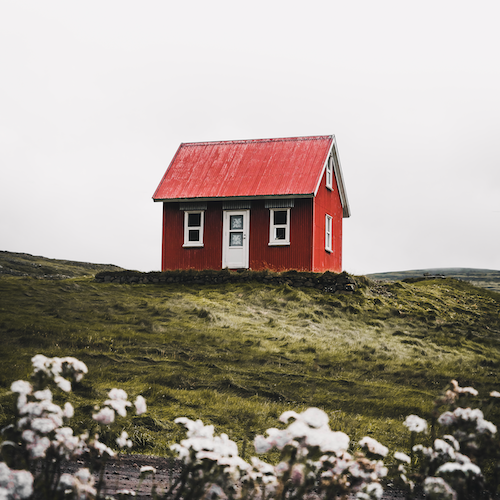} & \includegraphics[width=0.17\textwidth]{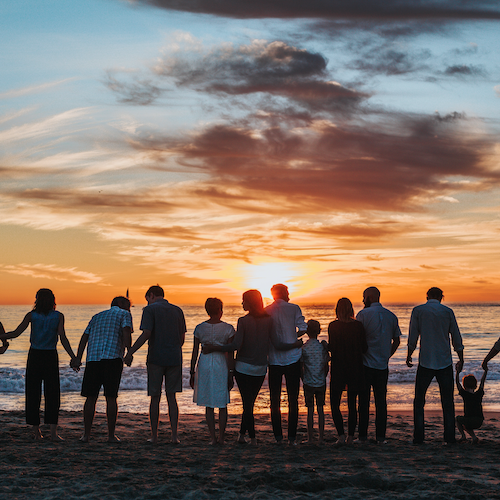}& \includegraphics[width=0.17\textwidth]{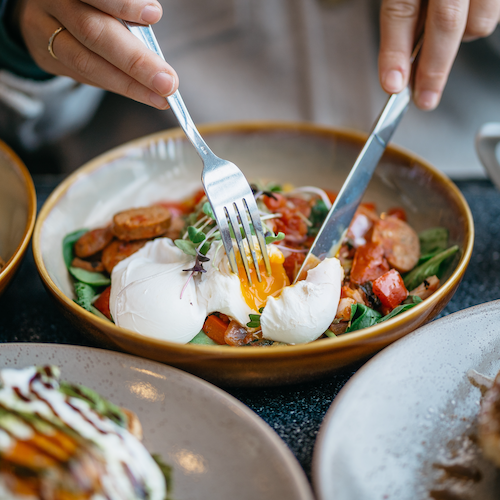}\\
\textcolor{completioncolor}{the flag was removed from the building after the fire broke out} & \textcolor{completioncolor}{: and beverage type at the beach. here are some of the most beautiful things i have ever seen.} &  \textcolor{completioncolor}{: the food is good for the body! :) - a-baked chicken. person. \~ photo by author} \\
\end{tabular}
\caption{JAT image captioning examples. The theme is usually correct, although the relevance is sometimes limited.}
\label{fig:image_captioning}
\vskip -0.1in
\end{figure}

We present a showcase of JAT's capabilities, with a particular focus on text completion and image captioning.
It's important to note that JAT is much smaller and has a much lower training budget than the specialized models for these tasks. 
Therefore, instead of comparing it to these expert systems, we want to demonstrate its intrinsic capabilities. Figure \ref{fig:image_captioning} shows a selection of captioning results to illustrate how the model interprets and describes visual data. 
Additional examples are given in Appendix \ref{appendix:image_captioning}.
Figure \ref{fig:text_completion_demo}, meanwhile, shows a series of text-based interactions that provide insight into its ability to complete text prompts.
These examples were chosen to highlight the model's basic capabilities in these areas, providing a realistic view of its current state of development and potential for future enhancements. In addition, we provide a demos\footnote{\url{https://huggingface.co/spaces/jat-project/text-completion}} for direct interaction and experimentation, allowing users to experience its functionalities.

\begin{figure}[ht]
\begin{center}
\begin{footnotesize}
\begin{tabular}{rp{9.6cm}}
\toprule
\textsc{Input} & \textsc{Model completion}  \\
\midrule
The weather today is & \textcolor{completioncolor}{a great time to the city of New York City. The city is a great place to stay in.} \\ \addlinespace
In the future, cars will & \textcolor{completioncolor}{be able to drive cars to the market. The new car will be built in the new market for the new car.} \\ \addlinespace
My favorite book is & \textcolor{completioncolor}{a book by the author of the book.} \\
\bottomrule
\end{tabular}
\end{footnotesize}
\end{center}
\caption{JAT text completion examples. The syntax is generally correct, the completion is on-topic, although the generated text may be wrong.}
\label{fig:text_completion_demo}
\end{figure}

\subsection{Sequential Decision-Making Tasks}

\begin{figure}[ht]
\begin{center}
\centerline{\includegraphics[width=0.6\textwidth]{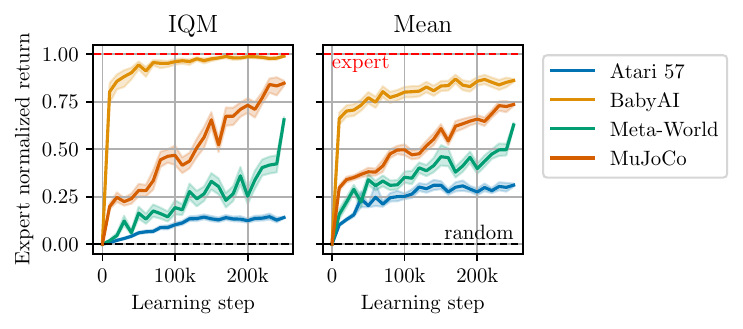}}
\caption{Aggregated expert normalized scores with 95\% Confidence Intervals (CIs) for each RL domain as a function of learning step.}
\label{fig:score_steps}
\end{center}
\vskip -0.2in
\end{figure}

\begin{figure*}[ht]
\begin{center}
\centerline{\includegraphics[width=0.99\textwidth]{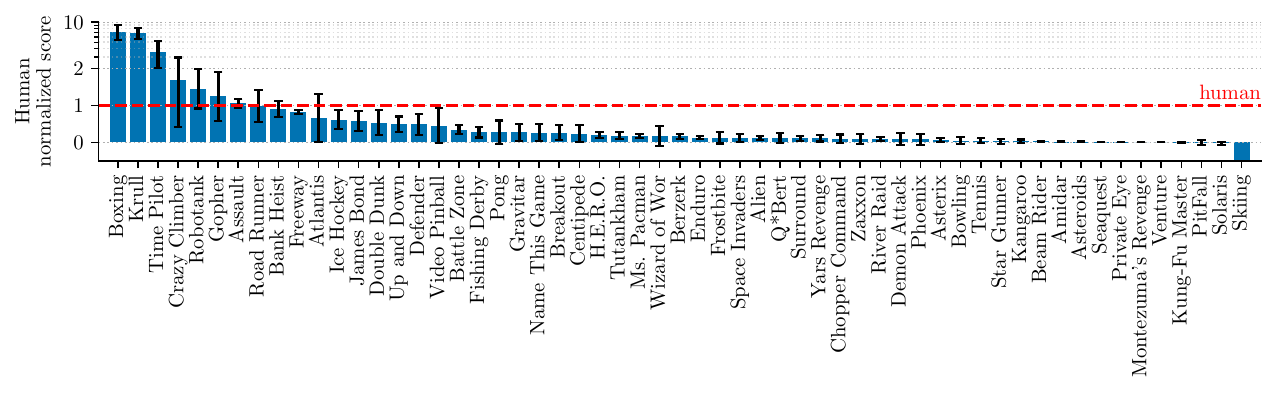}}
\caption{Human normalized scores for the JAT agent on the Atari 57 benchmark.}
\label{fig:atari_normalized}
\end{center}
\vskip -0.2in
\end{figure*}

We save checkpoints regularly during training. We evaluate each checkpoint on all the tasks on which it has been trained. Unlike Gato, the evaluation does not require any data to be used as a prompt. We show empirically in Appendix \ref{appendix:reward_task_determinant} that despite the absence of a prompt, and even in the worst case of our study, the agent still manages to identify the requested task.
For each task, we collect 10 evaluation episodes and normalize by the average expert score of the dataset for this task. For the final checkpoint, we use 100 evaluation episodes. We then aggregate the results by domain.
Figure \ref{fig:score_steps} shows the evolution of the aggregate score for each domain during learning, and Figure \ref{fig:atari_normalized} focuses on Atari, showing the human normalized score for each environment.
The final results are presented in detail in Appendix \ref{appendix:full_results}.

The final agent achieves a average expert normalized interquartile mean (IQM) of 65.8\%, demonstrating the network's ability to effectively mimic expert agents across a wide range of tasks. The agent achieves 14.1\% of the expert's score on the Atari 57 benchmark, corresponding to 37.6\% of human performance, and exceeding the average human level in 21 games. For the BabyAI benchmark, JAT achieves a normalized score of 99.0\%. This score falls below 50\% for only one task, namely Move Two Across S8N9. 
For this benchmark, however, there is no guarantee that the expert score can be achieved, since the bot used for dataset collection has access to the full state of the environment, while the interacting agents only have access to a partial observation.
Finally, in the MuJoCo and Meta-World, JAT records scores of 84.8\% and 65.5\%, respectively.
Although JAT reaches expert level for a fair number of Meta-World tasks, we note that some, such as basketball, have not been learned at all. Insofar as the action and observation spaces are identical for all tasks in this benchmark, these failures may be due to task indeterminacy, which we explore in more detail in Appendix \ref{appendix:reward_task_determinant}. Future research will have to confirm this hypothesis.
We also note that some domains are mastered more quickly than others; in particular, BabyAI achieves a score of 90\% after only 30,000 learning steps. We hypothesize that the high semantic similarity of the tasks enables a strong positive transfer, without however providing any proof of this. The Appendix \ref{appendix:full_results} presents the final results in detail.

Although the results achieved are commendable, for a fair comparison we limit our benchmarking to Gato only, as it is the only truly comparable baseline.
\citet{reed2022generalist} present results only for the 1.18 billion parameter version of Gato, which is 6 times larger than JAT. Its results are normalized to expert performance. Since we don't have access to the normalization parameters, we estimated scores for the random agents, which may not be exactly the same as those used by \citet{reed2022generalist}, and used our expert scores for normalization, even though they obviously do not match those used by \citet{reed2022generalist}. Therefore, comparisons of these normalized scores should be interpreted with great caution.
On the Atari benchmark, JAT achieves an average normalized score of 31.1\% outperforming Gato, which reports a score of 30.9\%.
For BabyAI, JAT achieves an average normalized score of 86.2\%, close to the Gato score of 93.2\%. Our study, however, is made with 39 tasks versus the 46 used in Gato's training, with the specific seven additional tasks in their study remaining unidentified. 
Since our evaluation includes all of the hardest tasks mentioned in their study, the seven missing tasks are likely to be easier, suggesting a harder test scenario in our study.
For Meta-World, JAT achieves an average normalized score of 62.8\%, which is below the 87.0\% reported for Gato.
On the MuJoCo benchmark, JAT achieves an average normalized score of 73.6\%. While Gato's training doesn't use MuJoCo, it's worth noting that they use the DMC benchmark, which shares some similarities. For reference, on the DMC benchmark \citep{tassa2018deepmind}, Gato achieves an average score of 63.6\%.

\subsection{Predicting the Observations Does Help}
\label{subsec:ablation_obs_loss}

The model's main task is to predict the actions that maximize the sum of future rewards. Its ability to predict future observations is therefore not the main concern. However, can this ability contribute to better prediction of actions or accelerate the learning process? Two contrasting hypotheses emerge: firstly, learning to predict observations could serve as an auxiliary objective, directing the learning process towards a deeper understanding of the environment, which could lead to improved and faster learning. Conversely, this prediction learning could serve as a distracting objective: instead of excelling in action prediction, the model might only achieve moderate performance in both action and observation prediction. This could slow down the learning process, resulting in a lower overall performance score. \citet{reed2022generalist} choose not to predict the observation, but does not study the influence of this prediction on learning.

To answer this question, we use a loss function that combines observation loss ($\mathcal{L}_\mathrm{obs}$) and action loss ($\mathcal{L}_\mathrm{act}$), balanced by a weighting parameter $\kappa$. The function is defined as:
\begin{equation}
\mathcal{L} = \kappa \cdot \mathcal{L}_\mathrm{obs} + (1 - \kappa) \cdot \mathcal{L}_\mathrm{act}
\end{equation}

\begin{figure}[ht]
\begin{center}
\centerline{\includegraphics[width=0.6\textwidth]{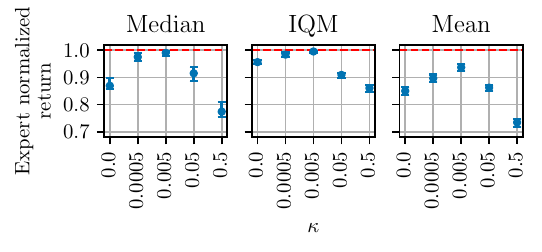}}
\caption{Aggregate measures with 95\% CIs for the study on the influence of observation prediction learning for selected tasks. The results presented cover the selected range of $\kappa$ values and are based on 100 evaluations per task. Optimal $\kappa$ selection can significantly improve agent performance.}
\label{fig:kappa_aggregated}
\end{center}
\vskip -0.2in
\end{figure}

We select a range of values for $\kappa$ and train the model on a subset of 6 dataset tasks from different domains (Freeway, Pong, ButtonPressWall, WindowClose, Ant and DoubleInvertedPendulum). 
Figure \ref{fig:kappa_aggregated} compares the results at the end of training for the different values of $\kappa$.

In our study of the $\kappa$ coefficient and its impact on learning, we find an interesting balance. When set to the highest value in our range ($\kappa=0.5$), the learning process seems to be somewhat hindered by the additional objective. On the other hand, at lower $\kappa$ values, this added task of predicting observations doesn't significantly impact learning, leading to scores that are similar to the base score of $94.5 \pm 1.1\%$, which we get when predicting observations isn't part of the objective. The sweet spot appears to be around $\kappa=0.005$. Learning to predict observations doesn't distract but actually improves the agent's learning efficiency, achieving an near-optimal score of $99.1 \pm 0.4\%$. This finding highlights that adding observation prediction into the learning process is beneficial, provided it's balanced correctly.

\section{Conclusion}

In this study, we introduce JAT, a novel multi-modal framework for general RL agents. JAT features the ability to handle diverse tasks of varying complexity using a single set of parameters. Its innovations include a new transformer-based structure that efficiently addresses sequential decision-making, CV, and NLP tasks. We also show that joint learning of observation prediction significantly improves performance in sequential decision-making tasks. We've open-sourced our training dataset, which includes a wide range of sequential decision-making data as well as extensive language and visual data. We believe that JAT represents an important and valuable step towards general-purpose RL models.

This study reveals several avenues for improvement. A primary challenge is the joint learning of tasks characterized by high heterogeneity. Our dataset features variations in size, task complexity, and accuracy requirements for optimal performance. Our current approach, which uses basic sample and loss weighting, partially addresses this challenge. A refinement of task sampling could potentially account for task difficulty, although quantifying \textit{difficulty} remains a challenge.
Another important challenge is imitation learning. While our current method relies on rudimentary behavioral cloning, the use of more advanced imitation learning techniques is likely to yield better results.
In addition, improving the quality of expert data is a clear opportunity for improvement. For example, in the Asterix task, our model's expert score (3699.6) lags significantly behind the scores achieved by agents such as R2D2 (999,153.3) \citep{kapturowski2019recurrent}. Using the best RL agent for each specific task could significantly improve the overall scores in our dataset, leading to better results when distilled for the generalist agent.

\section*{Broader Impact}

What sets generalist agents apart is their ability to produce output in a wide range of modalities (textual, visual, virtual or physical control) for multiple applications. This versatility introduces a potential for cross-domain generalisation, although no work to our knowledge actually demonstrates this transfer. The theoretical risk is that such agents transpose behaviours from one domain to another in an inappropriate or undesirable way, and raises important ethical and safety questions. For example, translating aggressive actions in a virtual environment (which may be legitimate in the context of a game or film) into harmful behaviour in the real world. While the alignment of large language models (LLMs) with human values and preferences has been extensively studied \citep{ouyang2022training,rafailov2023direct, azar2023general}, the application of these alignment strategies to the broader category of generalist models has not been examined in depth. The adaptation of existing alignment methodologies to the requirements of generalist agents is necessary to enable more reliable and secure models.

\begin{ack}
    
This work was granted access to the HPC resources of IDRIS under the allocation 2022-[AD011012172R1] made by GENCI.
    
\end{ack}



\bibliography{references}

\begin{thebibliography}{53}
\providecommand{\natexlab}[1]{#1}
\providecommand{\url}[1]{\texttt{#1}}
\expandafter\ifx\csname urlstyle\endcsname\relax
  \providecommand{\doi}[1]{doi: #1}\else
  \providecommand{\doi}{doi: \begingroup \urlstyle{rm}\Url}\fi

\bibitem[Agarwal et~al.(2021)Agarwal, Schwarzer, Castro, Courville, and Bellemare]{agarwal2021deep}
Agarwal, R., Schwarzer, M., Castro, P.~S., Courville, A.~C., and Bellemare, M.~G.
\newblock {Deep Reinforcement Learning at the Edge of the Statistical Precipice}.
\newblock In Ranzato, M., Beygelzimer, A., Dauphin, Y.~N., Liang, P., and Vaughan, J.~W. (eds.), \emph{Advances in Neural Information Processing Systems 34: Annual Conference on Neural Information Processing Systems 2021, NeurIPS 2021, December 6-14, 2021, virtual}, pp.\  29304--29320, 2021.
\newblock URL \url{https://proceedings.neurips.cc/paper/2021/hash/f514cec81cb148559cf475e7426eed5e-Abstract.html}.

\bibitem[Alayrac et~al.(2022)Alayrac, Donahue, Luc, Miech, Barr, Hasson, Lenc, Mensch, Millican, Reynolds, Ring, Rutherford, Cabi, Han, Gong, Samangooei, Monteiro, Menick, Borgeaud, Brock, Nematzadeh, Sharifzadeh, Binkowski, Barreira, Vinyals, Zisserman, and Simonyan]{alayrac2022flamingo}
Alayrac, J., Donahue, J., Luc, P., Miech, A., Barr, I., Hasson, Y., Lenc, K., Mensch, A., Millican, K., Reynolds, M., Ring, R., Rutherford, E., Cabi, S., Han, T., Gong, Z., Samangooei, S., Monteiro, M., Menick, J.~L., Borgeaud, S., Brock, A., Nematzadeh, A., Sharifzadeh, S., Binkowski, M., Barreira, R., Vinyals, O., Zisserman, A., and Simonyan, K.
\newblock {Flamingo: a Visual Language Model for Few-Shot Learning}.
\newblock In Koyejo, S., Mohamed, S., Agarwal, A., Belgrave, D., Cho, K., and Oh, A. (eds.), \emph{Advances in Neural Information Processing Systems 35: Annual Conference on Neural Information Processing Systems 2022, NeurIPS 2022, New Orleans, LA, USA, November 28 - December 9, 2022}, 2022.
\newblock URL \url{http://papers.nips.cc/paper\_files/paper/2022/hash/960a172bc7fbf0177ccccbb411a7d800-Abstract-Conference.html}.

\bibitem[Azar et~al.(2023)Azar, Rowland, Piot, Guo, Calandriello, Valko, and Munos]{azar2023general}
Azar, M.~G., Rowland, M., Piot, B., Guo, D., Calandriello, D., Valko, M., and Munos, R.
\newblock {A General Theoretical Paradigm to Understand Learning from Human Preferences}.
\newblock \emph{arXiv preprint arXiv:2310.12036}, 2023.

\bibitem[Bellemare et~al.(2013)Bellemare, Naddaf, Veness, and Bowling]{bellemare2013arcade}
Bellemare, M.~G., Naddaf, Y., Veness, J., and Bowling, M.
\newblock {The Arcade Learning Environment: An Evaluation Platform for General Agents}.
\newblock \emph{Journal of Artificial Intelligence Research}, 47:\penalty0 253--279, 2013.

\bibitem[Beltagy et~al.(2020)Beltagy, Peters, and Cohan]{beltagy2020longformer}
Beltagy, I., Peters, M.~E., and Cohan, A.
\newblock {Longformer: The Long-Document Transformer}.
\newblock \emph{arXiv preprint arXiv:2004.05150}, 2020.

\bibitem[Black et~al.(2021)Black, Leo, Wang, Leahy, and Biderman]{black2021gpt}
Black, S., Leo, G., Wang, P., Leahy, C., and Biderman, S.
\newblock {GPT-Neo: Large Scale Autoregressive Language Modeling with Mesh-Tensorflow}, March 2021.

\bibitem[Brockman et~al.(2016)Brockman, Cheung, Pettersson, Schneider, Schulman, Tang, and Zaremba]{brockman2016openai}
Brockman, G., Cheung, V., Pettersson, L., Schneider, J., Schulman, J., Tang, J., and Zaremba, W.
\newblock {OpenAI Gym}.
\newblock \emph{arXiv preprint arXiv:1606.01540}, 2016.

\bibitem[Brohan et~al.(2023{\natexlab{a}})Brohan, Brown, Carbajal, Chebotar, Chen, Choromanski, Ding, Driess, Dubey, Finn, Florence, Fu, Arenas, Gopalakrishnan, Han, Hausman, Herzog, Hsu, Ichter, Irpan, Joshi, Julian, Kalashnikov, Kuang, Leal, Lee, Lee, Levine, Lu, Michalewski, Mordatch, Pertsch, Rao, Reymann, Ryoo, Salazar, Sanketi, Sermanet, Singh, Singh, Soricut, Tran, Vanhoucke, Vuong, Wahid, Welker, Wohlhart, Wu, Xia, Xiao, Xu, Xu, Yu, and Zitkovich]{brohan2023rt2}
Brohan, A., Brown, N., Carbajal, J., Chebotar, Y., Chen, X., Choromanski, K., Ding, T., Driess, D., Dubey, A., Finn, C., Florence, P., Fu, C., Arenas, M.~G., Gopalakrishnan, K., Han, K., Hausman, K., Herzog, A., Hsu, J., Ichter, B., Irpan, A., Joshi, N.~J., Julian, R., Kalashnikov, D., Kuang, Y., Leal, I., Lee, L., Lee, T.~E., Levine, S., Lu, Y., Michalewski, H., Mordatch, I., Pertsch, K., Rao, K., Reymann, K., Ryoo, M.~S., Salazar, G., Sanketi, P., Sermanet, P., Singh, J., Singh, A., Soricut, R., Tran, H.~T., Vanhoucke, V., Vuong, Q., Wahid, A., Welker, S., Wohlhart, P., Wu, J., Xia, F., Xiao, T., Xu, P., Xu, S., Yu, T., and Zitkovich, B.
\newblock {RT-2: Vision-Language-Action Models Transfer Web Knowledge to Robotic Control}.
\newblock \emph{arXiv preprint arXiv:2307.15818}, 2023{\natexlab{a}}.

\bibitem[Brohan et~al.(2023{\natexlab{b}})Brohan, Brown, Carbajal, Chebotar, Dabis, Finn, Gopalakrishnan, Hausman, Herzog, Hsu, Ibarz, Ichter, Irpan, Jackson, Jesmonth, Joshi, Julian, Kalashnikov, Kuang, Leal, Lee, Levine, Lu, Malla, Manjunath, Mordatch, Nachum, Parada, Peralta, Perez, Pertsch, Quiambao, Rao, Ryoo, Salazar, Sanketi, Sayed, Singh, Sontakke, Stone, Tan, Tran, Vanhoucke, Vega, Vuong, Xia, Xiao, Xu, Xu, Yu, and Zitkovich]{brohan2023rt1}
Brohan, A., Brown, N., Carbajal, J., Chebotar, Y., Dabis, J., Finn, C., Gopalakrishnan, K., Hausman, K., Herzog, A., Hsu, J., Ibarz, J., Ichter, B., Irpan, A., Jackson, T., Jesmonth, S., Joshi, N.~J., Julian, R., Kalashnikov, D., Kuang, Y., Leal, I., Lee, K., Levine, S., Lu, Y., Malla, U., Manjunath, D., Mordatch, I., Nachum, O., Parada, C., Peralta, J., Perez, E., Pertsch, K., Quiambao, J., Rao, K., Ryoo, M.~S., Salazar, G., Sanketi, P.~R., Sayed, K., Singh, J., Sontakke, S., Stone, A., Tan, C., Tran, H.~T., Vanhoucke, V., Vega, S., Vuong, Q., Xia, F., Xiao, T., Xu, P., Xu, S., Yu, T., and Zitkovich, B.
\newblock {RT-1: Robotics Transformer for Real-World Control at Scale}.
\newblock In Bekris, K.~E., Hauser, K., Herbert, S.~L., and Yu, J. (eds.), \emph{Robotics: Science and Systems XIX, Daegu, Republic of Korea, July 10-14, 2023}, 2023{\natexlab{b}}.
\newblock \doi{10.15607/RSS.2023.XIX.025}.
\newblock URL \url{https://doi.org/10.15607/RSS.2023.XIX.025}.

\bibitem[Brown et~al.(2020)Brown, Mann, Ryder, Subbiah, Kaplan, Dhariwal, Neelakantan, Shyam, Sastry, Askell, Agarwal, Herbert{-}Voss, Krueger, Henighan, Child, Ramesh, Ziegler, Wu, Winter, Hesse, Chen, Sigler, Litwin, Gray, Chess, Clark, Berner, McCandlish, Radford, Sutskever, and Amodei]{brown2020language}
Brown, T.~B., Mann, B., Ryder, N., Subbiah, M., Kaplan, J., Dhariwal, P., Neelakantan, A., Shyam, P., Sastry, G., Askell, A., Agarwal, S., Herbert{-}Voss, A., Krueger, G., Henighan, T., Child, R., Ramesh, A., Ziegler, D.~M., Wu, J., Winter, C., Hesse, C., Chen, M., Sigler, E., Litwin, M., Gray, S., Chess, B., Clark, J., Berner, C., McCandlish, S., Radford, A., Sutskever, I., and Amodei, D.
\newblock {Language Models are Few-Shot Learners}.
\newblock In Larochelle, H., Ranzato, M., Hadsell, R., Balcan, M., and Lin, H. (eds.), \emph{Advances in Neural Information Processing Systems 33: Annual Conference on Neural Information Processing Systems 2020, NeurIPS 2020, December 6-12, 2020, virtual}, 2020.
\newblock URL \url{https://proceedings.neurips.cc/paper/2020/hash/1457c0d6bfcb4967418bfb8ac142f64a-Abstract.html}.

\bibitem[Chebotar et~al.(2023)Chebotar, Vuong, Hausman, Xia, Lu, Irpan, Kumar, Yu, Herzog, Pertsch, Gopalakrishnan, Ibarz, Nachum, Sontakke, Salazar, Tran, Peralta, Tan, Manjunath, Singh, Zitkovich, Jackson, Rao, Finn, and Levine]{chebotar2023q}
Chebotar, Y., Vuong, Q., Hausman, K., Xia, F., Lu, Y., Irpan, A., Kumar, A., Yu, T., Herzog, A., Pertsch, K., Gopalakrishnan, K., Ibarz, J., Nachum, O., Sontakke, S.~A., Salazar, G., Tran, H.~T., Peralta, J., Tan, C., Manjunath, D., Singh, J., Zitkovich, B., Jackson, T., Rao, K., Finn, C., and Levine, S.
\newblock {Q-Transformer: Scalable Offline Reinforcement Learning via Autoregressive Q-Functions}.
\newblock In Tan, J., Toussaint, M., and Darvish, K. (eds.), \emph{Conference on Robot Learning, CoRL 2023, 6-9 November 2023, Atlanta, GA, {USA}}, volume 229 of \emph{Proceedings of Machine Learning Research}, pp.\  3909--3928. {PMLR}, 2023.
\newblock URL \url{https://proceedings.mlr.press/v229/chebotar23a.html}.

\bibitem[Chen et~al.(2021)Chen, Lu, Rajeswaran, Lee, Grover, Laskin, Abbeel, Srinivas, and Mordatch]{chen2021decision}
Chen, L., Lu, K., Rajeswaran, A., Lee, K., Grover, A., Laskin, M., Abbeel, P., Srinivas, A., and Mordatch, I.
\newblock {Decision Transformer: Reinforcement Learning via Sequence Modeling}.
\newblock In Ranzato, M., Beygelzimer, A., Dauphin, Y.~N., Liang, P., and Vaughan, J.~W. (eds.), \emph{Advances in Neural Information Processing Systems 34: Annual Conference on Neural Information Processing Systems 2021, NeurIPS 2021, December 6-14, 2021, virtual}, pp.\  15084--15097, 2021.
\newblock URL \url{https://proceedings.neurips.cc/paper/2021/hash/7f489f642a0ddb10272b5c31057f0663-Abstract.html}.

\bibitem[Chen et~al.(2023)Chen, Wang, Changpinyo, Piergiovanni, Padlewski, Salz, Goodman, Grycner, Mustafa, Beyer, Kolesnikov, Puigcerver, Ding, Rong, Akbari, Mishra, Xue, Thapliyal, Bradbury, and Kuo]{chen2023pali}
Chen, X., Wang, X., Changpinyo, S., Piergiovanni, A.~J., Padlewski, P., Salz, D., Goodman, S., Grycner, A., Mustafa, B., Beyer, L., Kolesnikov, A., Puigcerver, J., Ding, N., Rong, K., Akbari, H., Mishra, G., Xue, L., Thapliyal, A.~V., Bradbury, J., and Kuo, W.
\newblock {PaLI: A Jointly-Scaled Multilingual Language-Image Model}.
\newblock In \emph{The Eleventh International Conference on Learning Representations, {ICLR} 2023, Kigali, Rwanda, May 1-5, 2023}. OpenReview.net, 2023.
\newblock URL \url{https://openreview.net/pdf?id=mWVoBz4W0u}.

\bibitem[Chevalier{-}Boisvert et~al.(2019)Chevalier{-}Boisvert, Bahdanau, Lahlou, Willems, Saharia, Nguyen, and Bengio]{chevalier_boisvert2019babyai}
Chevalier{-}Boisvert, M., Bahdanau, D., Lahlou, S., Willems, L., Saharia, C., Nguyen, T.~H., and Bengio, Y.
\newblock {BabyAI: A Platform to Study the Sample Efficiency of Grounded Language Learning}.
\newblock In \emph{7th International Conference on Learning Representations, {ICLR} 2019, New Orleans, LA, USA, May 6-9, 2019}. OpenReview.net, 2019.
\newblock URL \url{https://openreview.net/forum?id=rJeXCo0cYX}.

\bibitem[Chevalier-Boisvert et~al.(2023)Chevalier-Boisvert, Dai, Towers, de~Lazcano, Willems, Lahlou, Pal, Castro, and Terry]{chevalier_boisvert2023minigrid}
Chevalier-Boisvert, M., Dai, B., Towers, M., de~Lazcano, R., Willems, L., Lahlou, S., Pal, S., Castro, P.~S., and Terry, J.
\newblock {Minigrid \& Miniworld: Modular \& Customizable Reinforcement Learning Environments for Goal-Oriented Tasks}.
\newblock \emph{arXiv preprint arXiv:2306.13831}, 2023.

\bibitem[Dosovitskiy et~al.(2021)Dosovitskiy, Beyer, Kolesnikov, Weissenborn, Zhai, Unterthiner, Dehghani, Minderer, Heigold, Gelly, Uszkoreit, and Houlsby]{dosovitskiy2021image}
Dosovitskiy, A., Beyer, L., Kolesnikov, A., Weissenborn, D., Zhai, X., Unterthiner, T., Dehghani, M., Minderer, M., Heigold, G., Gelly, S., Uszkoreit, J., and Houlsby, N.
\newblock {An Image is Worth 16x16 Words: Transformers for Image Recognition at Scale}.
\newblock In \emph{9th International Conference on Learning Representations, {ICLR} 2021, Virtual Event, Austria, May 3-7, 2021}. OpenReview.net, 2021.

\bibitem[Driess et~al.(2023)Driess, Xia, Sajjadi, Lynch, Chowdhery, Ichter, Wahid, Tompson, Vuong, Yu, et~al.]{driess2023palm}
Driess, D., Xia, F., Sajjadi, M.~S., Lynch, C., Chowdhery, A., Ichter, B., Wahid, A., Tompson, J., Vuong, Q., Yu, T., et~al.
\newblock {PaLM-E: An Embodied Multimodal Language Model}.
\newblock \emph{arXiv preprint arXiv:2303.03378}, 2023.

\bibitem[Espeholt et~al.(2018)Espeholt, Soyer, Munos, Simonyan, Mnih, Ward, Doron, Firoiu, Harley, Dunning, Legg, and Kavukcuoglu]{espeholt2018impala}
Espeholt, L., Soyer, H., Munos, R., Simonyan, K., Mnih, V., Ward, T., Doron, Y., Firoiu, V., Harley, T., Dunning, I., Legg, S., and Kavukcuoglu, K.
\newblock {IMPALA: Scalable Distributed Deep-RL with Importance Weighted Actor-Learner Architectures}.
\newblock In Dy, J.~G. and Krause, A. (eds.), \emph{Proceedings of the 35th International Conference on Machine Learning, {ICML} 2018, Stockholmsm{\"{a}}ssan, Stockholm, Sweden, July 10-15, 2018}, volume~80 of \emph{Proceedings of Machine Learning Research}, pp.\  1406--1415. {PMLR}, 2018.
\newblock URL \url{http://proceedings.mlr.press/v80/espeholt18a.html}.

\bibitem[Foundation()]{wikidump}
Foundation, W.
\newblock {Wikimedia Downloads}.
\newblock URL \url{https://dumps.wikimedia.org}.

\bibitem[Gugger et~al.(2022)Gugger, Debut, Wolf, Schmid, Mueller, Mangrulkar, Sun, and Bossan]{gugger2022accelerate}
Gugger, S., Debut, L., Wolf, T., Schmid, P., Mueller, Z., Mangrulkar, S., Sun, M., and Bossan, B.
\newblock {Accelerate: Training and Inference at Scale Made Simple, Efficient and Adaptable}.
\newblock \url{https://github.com/huggingface/accelerate}, 2022.

\bibitem[Janner et~al.(2021)Janner, Li, and Levine]{janner2021offline}
Janner, M., Li, Q., and Levine, S.
\newblock {Offline Reinforcement Learning as One Big Sequence Modeling Problem}.
\newblock In Ranzato, M., Beygelzimer, A., Dauphin, Y.~N., Liang, P., and Vaughan, J.~W. (eds.), \emph{Advances in Neural Information Processing Systems 34: Annual Conference on Neural Information Processing Systems 2021, NeurIPS 2021, December 6-14, 2021, virtual}, pp.\  1273--1286, 2021.
\newblock URL \url{https://proceedings.neurips.cc/paper/2021/hash/099fe6b0b444c23836c4a5d07346082b-Abstract.html}.

\bibitem[Jiang et~al.(2022)Jiang, Gupta, Zhang, Wang, Dou, Chen, Fei-Fei, Anandkumar, Zhu, and Fan]{jiang2022vima}
Jiang, Y., Gupta, A., Zhang, Z., Wang, G., Dou, Y., Chen, Y., Fei-Fei, L., Anandkumar, A., Zhu, Y., and Fan, L.
\newblock {VIMA: General Robot Manipulation with Multimodal Prompts}.
\newblock \emph{arXiv preprint arXiv:2210.03094}, 2022.

\bibitem[Kapturowski et~al.(2019)Kapturowski, Ostrovski, Quan, Munos, and Dabney]{kapturowski2019recurrent}
Kapturowski, S., Ostrovski, G., Quan, J., Munos, R., and Dabney, W.
\newblock {Recurrent Experience Replay in Distributed Reinforcement Learning}.
\newblock In \emph{7th International Conference on Learning Representations, {ICLR} 2019, New Orleans, LA, USA, May 6-9, 2019}. OpenReview.net, 2019.
\newblock URL \url{https://openreview.net/forum?id=r1lyTjAqYX}.

\bibitem[Laurençon et~al.(2023{\natexlab{a}})Laurençon, Saulnier, Tronchon, Bekman, Singh, Lozhkov, Wang, Karamcheti, Rush, Kiela, Cord, and Sanh]{laurençon2023obelisc}
Laurençon, H., Saulnier, L., Tronchon, L., Bekman, S., Singh, A., Lozhkov, A., Wang, T., Karamcheti, S., Rush, A.~M., Kiela, D., Cord, M., and Sanh, V.
\newblock {OBELISC: An Open Web-Scale Filtered Dataset of Interleaved Image-Text Documents}.
\newblock \emph{arXiv preprint arXiv:2306.16527}, 2023{\natexlab{a}}.

\bibitem[Laurençon et~al.(2023{\natexlab{b}})Laurençon, Saulnier, Wang, Akiki, del Moral, Scao, Werra, Mou, Ponferrada, Nguyen, Frohberg, Šaško, Lhoest, McMillan-Major, Dupont, Biderman, Rogers, allal, Toni, Pistilli, Nguyen, Nikpoor, Masoud, Colombo, de~la Rosa, Villegas, Thrush, Longpre, Nagel, Weber, Muñoz, Zhu, Strien, Alyafeai, Almubarak, Vu, Gonzalez-Dios, Soroa, Lo, Dey, Suarez, Gokaslan, Bose, Adelani, Phan, Tran, Yu, Pai, Chim, Lepercq, Ilic, Mitchell, Luccioni, and Jernite]{laurençon2023bigscience}
Laurençon, H., Saulnier, L., Wang, T., Akiki, C., del Moral, A.~V., Scao, T.~L., Werra, L.~V., Mou, C., Ponferrada, E.~G., Nguyen, H., Frohberg, J., Šaško, M., Lhoest, Q., McMillan-Major, A., Dupont, G., Biderman, S., Rogers, A., allal, L.~B., Toni, F.~D., Pistilli, G., Nguyen, O., Nikpoor, S., Masoud, M., Colombo, P., de~la Rosa, J., Villegas, P., Thrush, T., Longpre, S., Nagel, S., Weber, L., Muñoz, M., Zhu, J., Strien, D.~V., Alyafeai, Z., Almubarak, K., Vu, M.~C., Gonzalez-Dios, I., Soroa, A., Lo, K., Dey, M., Suarez, P.~O., Gokaslan, A., Bose, S., Adelani, D., Phan, L., Tran, H., Yu, I., Pai, S., Chim, J., Lepercq, V., Ilic, S., Mitchell, M., Luccioni, S.~A., and Jernite, Y.
\newblock {The BigScience ROOTS Corpus: A 1.6TB Composite Multilingual Dataset}.
\newblock \emph{arXiv preprint arXiv:2303.03915}, 2023{\natexlab{b}}.

\bibitem[Lee et~al.(2022)Lee, Nachum, Yang, Lee, Freeman, Guadarrama, Fischer, Xu, Jang, Michalewski, and Mordatch]{lee2022multi}
Lee, K.-H., Nachum, O., Yang, M.~S., Lee, L., Freeman, D., Guadarrama, S., Fischer, I., Xu, W., Jang, E., Michalewski, H., and Mordatch, I.
\newblock {Multi-Game Decision Transformers}.
\newblock In Koyejo, S., Mohamed, S., Agarwal, A., Belgrave, D., Cho, K., and Oh, A. (eds.), \emph{Advances in Neural Information Processing Systems 35: Annual Conference on Neural Information Processing Systems 2022, NeurIPS 2022, New Orleans, LA, USA, November 28 - December 9, 2022}, volume~35, pp.\  27921--27936. Curran Associates, Inc., 2022.
\newblock URL \url{http://papers.nips.cc/paper\_files/paper/2022/hash/b2cac94f82928a85055987d9fd44753f-Abstract-Conference.html}.

\bibitem[Li et~al.(2023)Li, Luo, Lin, Zhang, Lu, and Ye]{li2023survey}
Li, W., Luo, H., Lin, Z., Zhang, C., Lu, Z., and Ye, D.
\newblock {A Survey on Transformers in Reinforcement Learning}.
\newblock \emph{Transactions on Machine Learning Research}, 2023.
\newblock ISSN 2835-8856.
\newblock URL \url{https://openreview.net/forum?id=r30yuDPvf2}.
\newblock Survey Certification.

\bibitem[Liu \& Abbeel(2023)Liu and Abbeel]{liu2023emergent}
Liu, H. and Abbeel, P.
\newblock {Emergent Agentic Transformer from Chain of Hindsight Experience}.
\newblock In Krause, A., Brunskill, E., Cho, K., Engelhardt, B., Sabato, S., and Scarlett, J. (eds.), \emph{International Conference on Machine Learning, {ICML} 2023, 23-29 July 2023, Honolulu, Hawaii, {USA}}, volume 202 of \emph{Proceedings of Machine Learning Research}, pp.\  21362--21374. {PMLR}, 2023.

\bibitem[Marino et~al.(2019)Marino, Rastegari, Farhadi, and Mottaghi]{marino2019ok}
Marino, K., Rastegari, M., Farhadi, A., and Mottaghi, R.
\newblock {OK-VQA: A Visual Question Answering Benchmark Requiring External Knowledge}.
\newblock In \emph{{IEEE} Conference on Computer Vision and Pattern Recognition, {CVPR} 2019, Long Beach, CA, USA, June 16-20, 2019}, pp.\  3195--3204. Computer Vision Foundation / {IEEE}, 2019.

\bibitem[Mnih et~al.(2015)Mnih, Kavukcuoglu, Silver, Rusu, Veness, Bellemare, Graves, Riedmiller, Fidjeland, Ostrovski, et~al.]{mnih2015human}
Mnih, V., Kavukcuoglu, K., Silver, D., Rusu, A.~A., Veness, J., Bellemare, M.~G., Graves, A., Riedmiller, M., Fidjeland, A.~K., Ostrovski, G., et~al.
\newblock {Human-Level Control Through Deep Reinforcement Learning}.
\newblock \emph{Nature}, 518\penalty0 (7540):\penalty0 529--533, 2015.

\bibitem[Moi \& Patry(2023)Moi and Patry]{moi2023huggingface}
Moi, A. and Patry, N.
\newblock {HuggingFace's Tokenizers}, April 2023.

\bibitem[Ortiz~Su{\'a}rez et~al.(2020)Ortiz~Su{\'a}rez, Romary, and Sagot]{ortiz_su_arez2020monolingual}
Ortiz~Su{\'a}rez, P.~J., Romary, L., and Sagot, B.
\newblock {A Monolingual Approach to Contextualized Word Embeddings for Mid-Resource Languages}.
\newblock In Jurafsky, D., Chai, J., Schluter, N., and Tetreault, J. (eds.), \emph{Proceedings of the 58th Annual Meeting of the Association for Computational Linguistics}, pp.\  1703--1714, Online, July 2020. Association for Computational Linguistics.
\newblock \doi{10.18653/v1/2020.acl-main.156}.
\newblock URL \url{https://aclanthology.org/2020.acl-main.156}.

\bibitem[Ouyang et~al.(2022)Ouyang, Wu, Jiang, Almeida, Wainwright, Mishkin, Zhang, Agarwal, Slama, Ray, Schulman, Hilton, Kelton, Miller, Simens, Askell, Welinder, Christiano, Leike, and Lowe]{ouyang2022training}
Ouyang, L., Wu, J., Jiang, X., Almeida, D., Wainwright, C.~L., Mishkin, P., Zhang, C., Agarwal, S., Slama, K., Ray, A., Schulman, J., Hilton, J., Kelton, F., Miller, L., Simens, M., Askell, A., Welinder, P., Christiano, P.~F., Leike, J., and Lowe, R.
\newblock {Training Language Models to Follow Instructions with Human Feedback}.
\newblock In Koyejo, S., Mohamed, S., Agarwal, A., Belgrave, D., Cho, K., and Oh, A. (eds.), \emph{Advances in Neural Information Processing Systems 35: Annual Conference on Neural Information Processing Systems 2022, NeurIPS 2022, New Orleans, LA, USA, November 28 - December 9, 2022}, 2022.
\newblock URL \url{http://papers.nips.cc/paper\_files/paper/2022/hash/b1efde53be364a73914f58805a001731-Abstract-Conference.html}.

\bibitem[Parisotto et~al.(2016)Parisotto, Ba, and Salakhutdinov]{parisotto2016actor}
Parisotto, E., Ba, L.~J., and Salakhutdinov, R.
\newblock {Actor-Mimic: Deep Multitask and Transfer Reinforcement Learning}.
\newblock In Bengio, Y. and LeCun, Y. (eds.), \emph{4th International Conference on Learning Representations, {ICLR} 2016, San Juan, Puerto Rico, May 2-4, 2016, Conference Track Proceedings}, 2016.
\newblock URL \url{http://arxiv.org/abs/1511.06342}.

\bibitem[Petrenko et~al.(2020)Petrenko, Huang, Kumar, Sukhatme, and Koltun]{petrenko2020sample}
Petrenko, A., Huang, Z., Kumar, T., Sukhatme, G.~S., and Koltun, V.
\newblock {Sample Factory: Egocentric 3D Control from Pixels at 100000 FPS with Asynchronous Reinforcement Learning}.
\newblock In \emph{Proceedings of the 37th International Conference on Machine Learning, {ICML} 2020, 13-18 July 2020, Virtual Event}, volume 119 of \emph{Proceedings of Machine Learning Research}, pp.\  7652--7662. {PMLR}, 2020.
\newblock URL \url{http://proceedings.mlr.press/v119/petrenko20a.html}.

\bibitem[Radford et~al.(2019)Radford, Wu, Child, Luan, Amodei, Sutskever, et~al.]{radford2019language}
Radford, A., Wu, J., Child, R., Luan, D., Amodei, D., Sutskever, I., et~al.
\newblock {Language Models are Unsupervised Multitask Learners}.
\newblock \emph{OpenAI blog}, 1\penalty0 (8):\penalty0 9, 2019.

\bibitem[Rafailov et~al.(2023)Rafailov, Sharma, Mitchell, Manning, Ermon, and Finn]{rafailov2023direct}
Rafailov, R., Sharma, A., Mitchell, E., Manning, C.~D., Ermon, S., and Finn, C.
\newblock {Direct Preference Optimization: Your Language Model is Secretly a Reward Model}.
\newblock In Oh, A., Naumann, T., Globerson, A., Saenko, K., Hardt, M., and Levine, S. (eds.), \emph{Advances in Neural Information Processing Systems 36: Annual Conference on Neural Information Processing Systems 2023, NeurIPS 2023, New Orleans, LA, USA, December 10 - 16, 2023}, 2023.
\newblock URL \url{http://papers.nips.cc/paper\_files/paper/2023/hash/a85b405ed65c6477a4fe8302b5e06ce7-Abstract-Conference.html}.

\bibitem[Raffel et~al.(2020)Raffel, Shazeer, Roberts, Lee, Narang, Matena, Zhou, Li, and Liu]{raffel2020exploring}
Raffel, C., Shazeer, N., Roberts, A., Lee, K., Narang, S., Matena, M., Zhou, Y., Li, W., and Liu, P.~J.
\newblock {Exploring the Limits of Transfer Learning with a Unified Text-to-Text Transformer}.
\newblock \emph{Journal of Machine Learning Research}, 21\penalty0 (140):\penalty0 1--67, 2020.
\newblock ISSN 1533-7928.
\newblock URL \url{http://jmlr.org/papers/v21/20-074.html}.

\bibitem[Reed et~al.(2022)Reed, Zolna, Parisotto, Colmenarejo, Novikov, Barth-maron, Gim{\'e}nez, Sulsky, Kay, Springenberg, Eccles, Bruce, Razavi, Edwards, Heess, Chen, Hadsell, Vinyals, Bordbar, and de~Freitas]{reed2022generalist}
Reed, S., Zolna, K., Parisotto, E., Colmenarejo, S.~G., Novikov, A., Barth-maron, G., Gim{\'e}nez, M., Sulsky, Y., Kay, J., Springenberg, J.~T., Eccles, T., Bruce, J., Razavi, A., Edwards, A., Heess, N., Chen, Y., Hadsell, R., Vinyals, O., Bordbar, M., and de~Freitas, N.
\newblock {A Generalist Agent}.
\newblock \emph{Transactions on Machine Learning Research}, 2022.
\newblock ISSN 2835-8856.
\newblock Featured Certification, Outstanding Certification.

\bibitem[Rusu et~al.(2016)Rusu, Colmenarejo, G{\"{u}}l{\c{c}}ehre, Desjardins, Kirkpatrick, Pascanu, Mnih, Kavukcuoglu, and Hadsell]{rusu2016policy}
Rusu, A.~A., Colmenarejo, S.~G., G{\"{u}}l{\c{c}}ehre, {\c{C}}., Desjardins, G., Kirkpatrick, J., Pascanu, R., Mnih, V., Kavukcuoglu, K., and Hadsell, R.
\newblock {Policy Distillation}.
\newblock In Bengio, Y. and LeCun, Y. (eds.), \emph{4th International Conference on Learning Representations, {ICLR} 2016, San Juan, Puerto Rico, May 2-4, 2016, Conference Track Proceedings}, 2016.
\newblock URL \url{http://arxiv.org/abs/1511.06295}.

\bibitem[Schulman et~al.(2017)Schulman, Wolski, Dhariwal, Radford, and Klimov]{schulman2017proximal}
Schulman, J., Wolski, F., Dhariwal, P., Radford, A., and Klimov, O.
\newblock {Proximal Policy Optimization Algorithms}.
\newblock \emph{arXiv preprint arXiv:1707.06347}, 2017.

\bibitem[Sennrich et~al.(2016)Sennrich, Haddow, and Birch]{sennrich2016neural}
Sennrich, R., Haddow, B., and Birch, A.
\newblock {Neural Machine Translation of Rare Words with Subword Units}.
\newblock In \emph{Proceedings of the 54th Annual Meeting of the Association for Computational Linguistics (Volume 1: Long Papers)}, pp.\  1715--1725, 2016.

\bibitem[Sharma et~al.(2018)Sharma, Ding, Goodman, and Soricut]{sharma2018conceptual}
Sharma, P., Ding, N., Goodman, S., and Soricut, R.
\newblock {Conceptual Captions: A Cleaned, Hypernymed, Image Alt-text Dataset For Automatic Image Captioning}.
\newblock In Gurevych, I. and Miyao, Y. (eds.), \emph{Proceedings of the 56th Annual Meeting of the Association for Computational Linguistics (Volume 1: Long Papers)}, pp.\  2556--2565, Melbourne, Australia, July 2018. Association for Computational Linguistics.

\bibitem[Song et~al.(2020)Song, Abdolmaleki, Springenberg, Clark, Soyer, Rae, Noury, Ahuja, Liu, Tirumala, Heess, Belov, Riedmiller, and Botvinick]{song2020v}
Song, H.~F., Abdolmaleki, A., Springenberg, J.~T., Clark, A., Soyer, H., Rae, J.~W., Noury, S., Ahuja, A., Liu, S., Tirumala, D., Heess, N., Belov, D., Riedmiller, M.~A., and Botvinick, M.~M.
\newblock {V-MPO: On-Policy Maximum a Posteriori Policy Optimization for Discrete and Continuous Control}.
\newblock In \emph{8th International Conference on Learning Representations, {ICLR} 2020, Addis Ababa, Ethiopia, April 26-30, 2020}. OpenReview.net, 2020.
\newblock URL \url{https://openreview.net/forum?id=SylOlp4FvH}.

\bibitem[Tassa et~al.(2018)Tassa, Doron, Muldal, Erez, Li, de~Las~Casas, Budden, Abdolmaleki, Merel, Lefrancq, Lillicrap, and Riedmiller]{tassa2018deepmind}
Tassa, Y., Doron, Y., Muldal, A., Erez, T., Li, Y., de~Las~Casas, D., Budden, D., Abdolmaleki, A., Merel, J., Lefrancq, A., Lillicrap, T.~P., and Riedmiller, M.~A.
\newblock {DeepMind Control Suite}.
\newblock \emph{arXiv preprint arXiv:1801.00690}, 2018.

\bibitem[Teh et~al.(2017)Teh, Bapst, Czarnecki, Quan, Kirkpatrick, Hadsell, Heess, and Pascanu]{teh2017distral}
Teh, Y.~W., Bapst, V., Czarnecki, W.~M., Quan, J., Kirkpatrick, J., Hadsell, R., Heess, N., and Pascanu, R.
\newblock {Distral: Robust multitask reinforcement learning}.
\newblock In Guyon, I., von Luxburg, U., Bengio, S., Wallach, H.~M., Fergus, R., Vishwanathan, S. V.~N., and Garnett, R. (eds.), \emph{Advances in Neural Information Processing Systems 30: Annual Conference on Neural Information Processing Systems 2017, December 4-9, 2017, Long Beach, CA, {USA}}, pp.\  4496--4506, 2017.
\newblock URL \url{https://proceedings.neurips.cc/paper/2017/hash/0abdc563a06105aee3c6136871c9f4d1-Abstract.html}.

\bibitem[Todorov et~al.(2012)Todorov, Erez, and Tassa]{todorov2012mujoco}
Todorov, E., Erez, T., and Tassa, Y.
\newblock {MuJoCo: A physics engine for model-based control}.
\newblock In \emph{2012 {IEEE/RSJ} International Conference on Intelligent Robots and Systems, {IROS} 2012, Vilamoura, Algarve, Portugal, October 7-12, 2012}, pp.\  5026--5033. {IEEE}, 2012.
\newblock \doi{10.1109/IROS.2012.6386109}.
\newblock URL \url{https://doi.org/10.1109/IROS.2012.6386109}.

\bibitem[Vaswani et~al.(2017)Vaswani, Shazeer, Parmar, Uszkoreit, Jones, Gomez, Kaiser, and Polosukhin]{vaswani2017attention}
Vaswani, A., Shazeer, N., Parmar, N., Uszkoreit, J., Jones, L., Gomez, A.~N., Kaiser, L., and Polosukhin, I.
\newblock {Attention is All you Need}.
\newblock In Guyon, I., von Luxburg, U., Bengio, S., Wallach, H.~M., Fergus, R., Vishwanathan, S. V.~N., and Garnett, R. (eds.), \emph{Advances in Neural Information Processing Systems 30: Annual Conference on Neural Information Processing Systems 2017, December 4-9, 2017, Long Beach, CA, {USA}}, pp.\  5998--6008, 2017.
\newblock URL \url{https://proceedings.neurips.cc/paper/2017/hash/3f5ee243547dee91fbd053c1c4a845aa-Abstract.html}.

\bibitem[Wolf et~al.(2020)Wolf, Debut, Sanh, Chaumond, Delangue, Moi, Cistac, Rault, Louf, Funtowicz, Davison, Shleifer, von Platen, Ma, Jernite, Plu, Xu, Scao, Gugger, Drame, Lhoest, and Rush]{wolf2020transformers}
Wolf, T., Debut, L., Sanh, V., Chaumond, J., Delangue, C., Moi, A., Cistac, P., Rault, T., Louf, R., Funtowicz, M., Davison, J., Shleifer, S., von Platen, P., Ma, C., Jernite, Y., Plu, J., Xu, C., Scao, T.~L., Gugger, S., Drame, M., Lhoest, Q., and Rush, A.~M.
\newblock {Transformers: State-of-the-Art Natural Language Processing}.
\newblock In \emph{Proceedings of the 2020 Conference on Empirical Methods in Natural Language Processing: System Demonstrations}, pp.\  38--45. Association for Computational Linguistics, October 2020.
\newblock URL \url{https://www.aclweb.org/anthology/2020.emnlp-demos.6}.

\bibitem[Yang et~al.(2020)Yang, Xu, Wu, and Wang]{yang2020multi}
Yang, R., Xu, H., Wu, Y., and Wang, X.
\newblock {Multi-Task Reinforcement Learning with Soft Modularization}.
\newblock In Larochelle, H., Ranzato, M., Hadsell, R., Balcan, M., and Lin, H. (eds.), \emph{Advances in Neural Information Processing Systems 33: Annual Conference on Neural Information Processing Systems 2020, NeurIPS 2020, December 6-12, 2020, virtual}, 2020.
\newblock URL \url{https://proceedings.neurips.cc/paper/2020/hash/32cfdce9631d8c7906e8e9d6e68b514b-Abstract.html}.

\bibitem[Yu et~al.(2019)Yu, Quillen, He, Julian, Hausman, Finn, and Levine]{yu2019meta}
Yu, T., Quillen, D., He, Z., Julian, R., Hausman, K., Finn, C., and Levine, S.
\newblock {Meta-World: A Benchmark and Evaluation for Multi-Task and Meta Reinforcement Learning}.
\newblock In Kaelbling, L.~P., Kragic, D., and Sugiura, K. (eds.), \emph{3rd Annual Conference on Robot Learning, CoRL 2019, Osaka, Japan, October 30 - November 1, 2019, Proceedings}, volume 100 of \emph{Proceedings of Machine Learning Research}, pp.\  1094--1100. {PMLR}, 2019.
\newblock URL \url{http://proceedings.mlr.press/v100/yu20a.html}.

\bibitem[Zeiler et~al.(2010)Zeiler, Krishnan, Taylor, and Fergus]{zeiler2010deconvolutional}
Zeiler, M.~D., Krishnan, D., Taylor, G.~W., and Fergus, R.
\newblock {Deconvolutional Networks}.
\newblock In \emph{The Twenty-Third {IEEE} Conference on Computer Vision and Pattern Recognition, {CVPR} 2010, San Francisco, CA, USA, 13-18 June 2010}, pp.\  2528--2535. {IEEE} Computer Society, 2010.
\newblock URL \url{https://doi.org/10.1109/CVPR.2010.5539957}.

\bibitem[Zheng et~al.(2022)Zheng, Zhang, and Grover]{zheng2022online}
Zheng, Q., Zhang, A., and Grover, A.
\newblock {Online Decision Transformer}.
\newblock \emph{arXiv preprint arXiv:2202.05607}, 2022.

\end{thebibliography}
\bibliographystyle{icml2024}

\newpage
\appendix
\onecolumn

\section{Full results}
\label{appendix:full_results}

\begin{figure}[p]
\centering
\includegraphics[width=\textwidth]{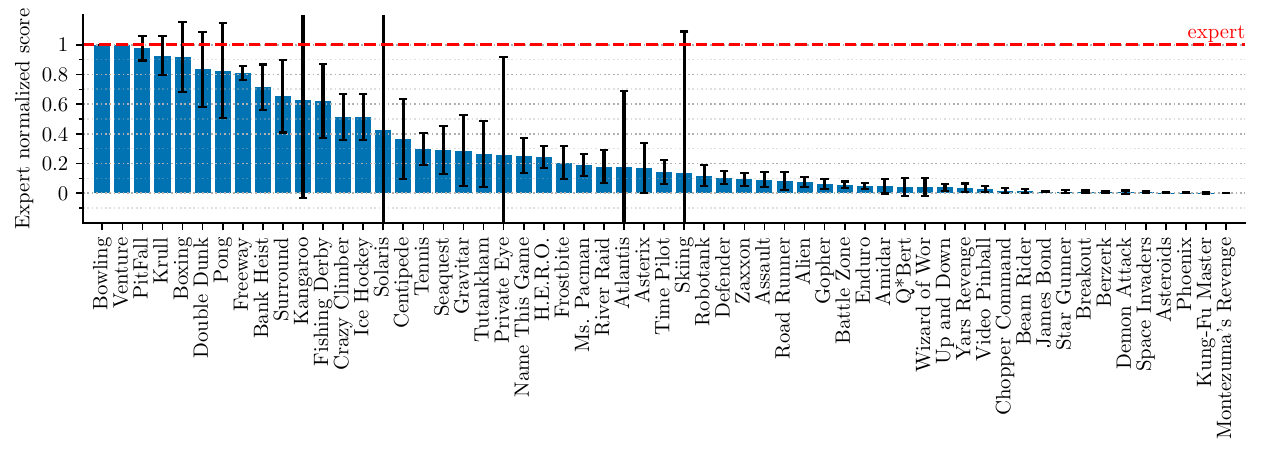}
\caption{Expert normalized episodic return for the JAT agent on the Atari 57 benchmark.}
\label{fig:comprehensive_scores_atari}
\end{figure}

\begin{figure}[p]
\centering
\includegraphics[width=\textwidth]{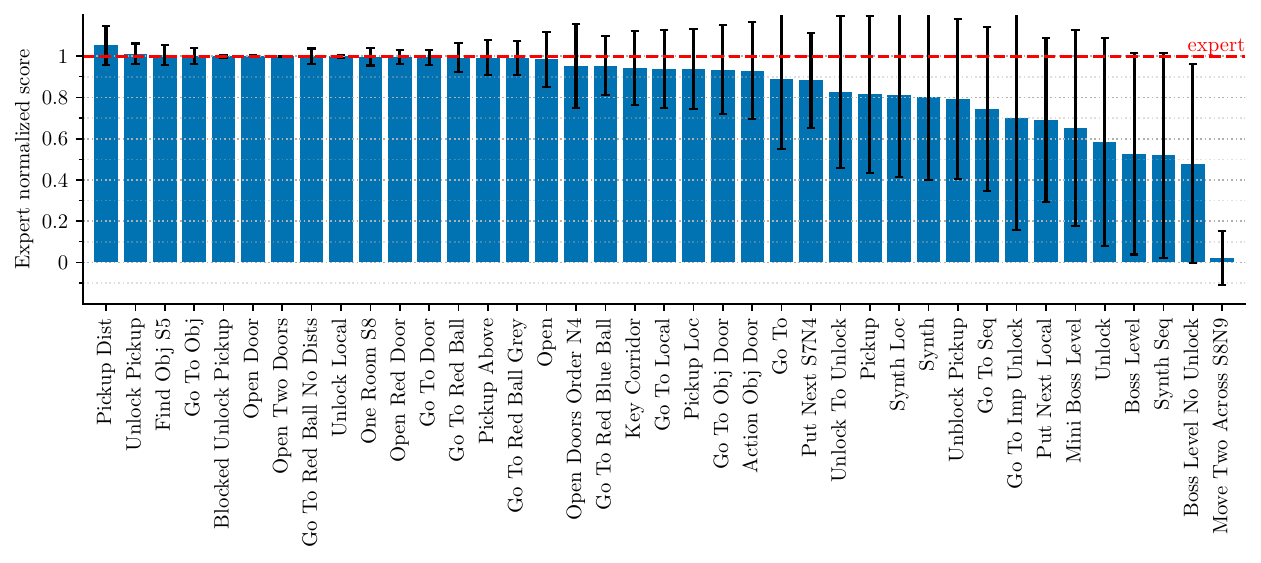}
\caption{Expert normalized episodic return for the JAT agent on the BabyAI benchmark.}
\label{fig:comprehensive_scores_babyai}
\end{figure}

\begin{figure}[p]
\centering
\includegraphics[width=\textwidth]{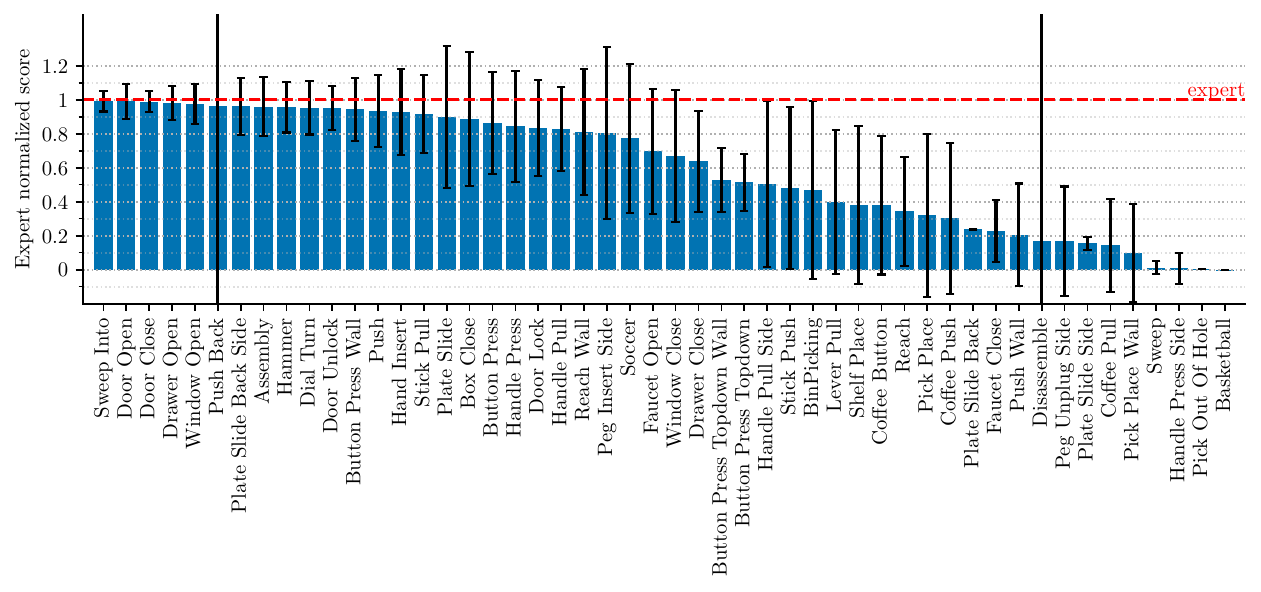}
\caption{Expert normalized episodic return for the JAT agent on the Meta-World benchmark.}
\label{fig:comprehensive_scores_metaworld}
\end{figure}

\begin{figure}[p]
\centering
\includegraphics[width=0.5\textwidth]{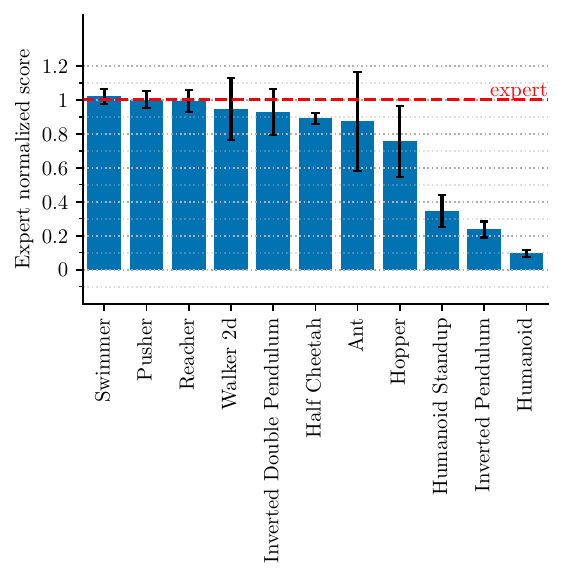}
\caption{Expert normalized episodic return for the JAT agent on the MuJoCo benchmark.}
\label{fig:comprehensive_scores_mujoco}
\end{figure}

This appendix contains a detailed view of the results of the trained JAT agent.
The score of the random agent for Atari games is sourced from \citep{mnih2015human}. In other domains, this score is approximated by averaging the returns from 1,000 episodes, where the agent selects actions uniformly across its action space.
The expert scores represent the average return in the dataset for the task. Meanwhile, the raw score is the average return achieved by the trained agent, based on 100 evaluation episodes. Both these scores, along with the trained agent, are accessible as open-source\footnote{\url{https://huggingface.co/jat-project/jat}}.
The normalized score is derived by comparing the agent's return to the expert's, calculated using the formula: $\frac{\text{{score}} - \text{{random\_score}}}{\text{{expert\_score}} - \text{random\_score}}$. It's important to note that in instances where the \textit{expert}, inaccurately named, does not fully master the task and thus scores similarly or lower than the random agent, the normalized score must be interpreted cautiously. Specifically, if this score falls below that of the random agent, as in the case of Bowling, normalization is not applied.
The results for Atari are presented in Table \ref{tab:comprehensive_scores_atari} and Figure \ref{fig:comprehensive_scores_atari},
for BabyAI in Table \ref{tab:comprehensive_scores_babyai} and Figure \ref{fig:comprehensive_scores_babyai},
for Meta-World in Table \ref{tab:comprehensive_scores_metaworld} and Figure \ref{fig:comprehensive_scores_metaworld}, and
for MuJoCo in Table \ref{tab:comprehensive_scores_mujoco} and Figure \ref{fig:comprehensive_scores_mujoco}.

\begin{table}[p]
\caption{Comparison of performance scores across tasks on Atari 57. The table presents the episodic return (score) achieved by a random agent (from \citep{mnih2015human}), scores of the expert agent (as averaged from the dataset), scores of the learned agent, and the expert normalized score calculated as $\frac{{\text{{score}} - \text{{random\_score}}}}{{\text{{expert\_score}} - \text{{random\_score}}}}$.}
\label{tab:comprehensive_scores_atari}
\fontsize{8pt}{8.5pt}\selectfont
\begin{center}
\begin{sc}
\begin{tabular}{lcccc}
\toprule
Task & Random agent & Expert & JAT (raw) & JAT (normalized) \\ \midrule
Alien & 227.8 & 16912.5 ± 7087.4 & 1474.9 ± 588.7 & 0.07 ± 0.04 \\ 
Amidar & 5.8 & 2164.7 ± 1229.5 & 104.9 ± 103.5 & 0.05 ± 0.05 \\ 
Assault & 222.4 & 15699.1 ± 9572.1 & 1650.1 ± 821.0 & 0.09 ± 0.05 \\ 
Asterix & 210.0 & 3699.6 ± 2421.3 & 800.0 ± 584.9 & 0.17 ± 0.17 \\ 
Asteroids & 719.0 & 177011.1 ± 35334.2 & 1385.3 ± 507.5 & 0.00 ± 0.00 \\ 
Atlantis & 12850.0 & 320679.6 ± 418247.4 & 66980.0 ± 158449.7 & 0.18 ± 0.51 \\ 
Bank Heist & 14.2 & 1322.4 ± 60.8 & 948.3 ± 199.9 & 0.71 ± 0.15 \\ 
Battle Zone & 236.0 & 295592.6 ± 161961.0 & 17420.0 ± 6071.5 & 0.06 ± 0.02 \\ 
Beam Rider & 363.9 & 29589.3 ± 16133.0 & 797.3 ± 328.3 & 0.01 ± 0.01 \\ 
Berzerk & 123.7 & 57085.3 ± 13104.5 & 687.3 ± 331.9 & 0.01 ± 0.01 \\ 
Bowling & 23.1 & 20.4 ± 7.3 & 22.4 ± 5.6 & N/A \\ 
Boxing & 0.1 & 98.0 ± 3.8 & 90.1 ± 23.0 & 0.92 ± 0.24 \\ 
Breakout & 1.7 & 703.0 ± 203.6 & 8.8 ± 5.6 & 0.01 ± 0.01 \\ 
Centipede & 2090.9 & 11624.3 ± 4918.3 & 5589.9 ± 2567.3 & 0.37 ± 0.27 \\ 
Chopper Command & 811.0 & 90990.6 ± 270876.9 & 2417.0 ± 1489.9 & 0.02 ± 0.02 \\ 
Crazy Climber & 10780.5 & 179296.9 ± 39862.1 & 97639.0 ± 26184.7 & 0.52 ± 0.16 \\ 
Defender & 2874.5 & 351958.3 ± 40466.8 & 39323.5 ± 15203.0 & 0.10 ± 0.04 \\ 
Demon Attack & 152.1 & 92195.2 ± 26174.8 & 815.3 ± 989.7 & 0.01 ± 0.01 \\ 
Double Dunk & -18.6 & 20.9 ± 3.6 & 14.4 ± 10.0 & 0.84 ± 0.25 \\ 
Enduro & 0.0 & 2292.2 ± 147.5 & 108.5 ± 42.7 & 0.05 ± 0.02 \\ 
Fishing Derby & -91.7 & 7.2 ± 25.1 & -30.4 ± 24.4 & 0.62 ± 0.25 \\ 
Freeway & 0.0 & 33.9 ± 0.3 & 27.5 ± 1.6 & 0.81 ± 0.05 \\ 
Frostbite & 65.2 & 13196.1 ± 4341.0 & 2769.6 ± 1445.6 & 0.21 ± 0.11 \\ 
Gopher & 257.6 & 81676.2 ± 46329.5 & 5340.6 ± 2547.1 & 0.06 ± 0.03 \\ 
Gravitar & 173.0 & 3986.6 ± 1729.0 & 1269.5 ± 903.0 & 0.29 ± 0.24 \\ 
H.E.R.O. & 1027.0 & 44677.4 ± 1754.4 & 11709.6 ± 3233.5 & 0.24 ± 0.07 \\ 
Ice Hockey & -11.2 & 25.2 ± 5.8 & 7.5 ± 5.6 & 0.51 ± 0.15 \\ 
James Bond & 29.0 & 27786.9 ± 33819.2 & 327.5 ± 123.2 & 0.01 ± 0.00 \\ 
Kangaroo & 52.0 & 574.0 ± 636.9 & 378.0 ± 344.0 & 0.62 ± 0.66 \\ 
Krull & 1598.0 & 11439.8 ± 1218.3 & 10720.5 ± 1284.1 & 0.93 ± 0.13 \\ 
Kung-Fu Master & 258.5 & 32392.8 ± 10006.6 & 288.0 ± 255.1 & 0.00 ± 0.01 \\ 
Montezuma's Revenge & 0.0 & 393.5 ± 50.4 & 0.0 ± 0.0 & 0.00 ± 0.00 \\ 
Ms. Pacman & 307.3 & 6896.1 ± 2032.0 & 1573.1 ± 484.0 & 0.19 ± 0.07 \\ 
Name This Game & 2292.3 & 22991.2 ± 2473.1 & 7523.3 ± 2471.4 & 0.25 ± 0.12 \\ 
Phoenix & 761.5 & 424583.2 ± 97649.2 & 2197.9 ± 1795.4 & 0.00 ± 0.00 \\ 
PitFall & -229.4 & -1.4 ± 4.5 & -6.7 ± 19.0 & 0.98 ± 0.08 \\ 
Pong & -20.7 & 21.0 ± 0.2 & 13.7 ± 13.3 & 0.82 ± 0.32 \\ 
Private Eye & 24.9 & 100.0 ± 0.0 & 44.0 ± 49.6 & 0.25 ± 0.66 \\ 
Q*Bert & 163.9 & 42971.4 ± 85070.7 & 1951.5 ± 2577.2 & 0.04 ± 0.06 \\ 
River Raid & 1338.5 & 14800.9 ± 7924.6 & 3758.5 ± 1536.7 & 0.18 ± 0.11 \\ 
Road Runner & 11.5 & 77942.8 ± 6088.6 & 6407.0 ± 4847.4 & 0.08 ± 0.06 \\ 
Robotank & 2.2 & 80.5 ± 13.3 & 11.3 ± 5.5 & 0.12 ± 0.07 \\ 
Seaquest & 68.4 & 2597.3 ± 386.1 & 804.0 ± 403.3 & 0.29 ± 0.16 \\ 
Skiing & -17098.0 & -10738.1 ± 111.1 & -16231.5 ± 6060.5 & 0.14 ± 0.95 \\ 
Solaris & 1236.3 & 1353.7 ± 517.0 & 1286.6 ± 446.7 & 0.43 ± 3.81 \\ 
Space Invaders & 148.0 & 29425.3 ± 23623.9 & 325.4 ± 163.4 & 0.01 ± 0.01 \\ 
Star Gunner & 664.0 & 360588.6 ± 49207.7 & 4379.0 ± 3027.2 & 0.01 ± 0.01 \\ 
Surround & -10.0 & 9.4 ± 0.8 & 2.7 ± 4.7 & 0.65 ± 0.24 \\ 
Tennis & -23.8 & 11.1 ± 7.6 & -13.5 ± 3.8 & 0.30 ± 0.11 \\ 
Time Pilot & 3568.0 & 69583.3 ± 29838.7 & 13028.0 ± 5222.6 & 0.14 ± 0.08 \\ 
Tutankham & 11.4 & 291.2 ± 30.4 & 85.7 ± 61.8 & 0.27 ± 0.22 \\ 
Up and Down & 533.4 & 429418.3 ± 7187.4 & 17768.7 ± 10322.0 & 0.04 ± 0.02 \\ 
Venture & 0.0 & 0.0 ± 0.0 & 0.0 ± 0.0 & N/A \\ 
Video Pinball & 0.0 & 441507.9 ± 283264.6 & 11917.4 ± 8204.3 & 0.03 ± 0.02 \\ 
Wizard of Wor & 563.5 & 49333.3 ± 16157.1 & 2544.0 ± 2902.4 & 0.04 ± 0.06 \\ 
Yars Revenge & 3092.9 & 270262.9 ± 161816.0 & 12532.7 ± 8062.8 & 0.04 ± 0.03 \\ 
Zaxxon & 32.5 & 73097.2 ± 14825.8 & 6902.0 ± 3206.1 & 0.09 ± 0.04 \\  \bottomrule
\end{tabular}
\end{sc}
\end{center}
\end{table}

\clearpage

\begin{table}[p]
\caption{Comparison of performance scores across tasks on BabyAI. The table presents the episodic return (score) achieved by a random agent (averaged over 1,000 episodes), scores of the expert agent (as averaged from the dataset), scores of the learned agent, and the expert normalized score calculated as $\frac{{\text{{score}} - \text{{random\_score}}}}{{\text{{expert\_score}} - \text{{random\_score}}}}$.}
\label{tab:comprehensive_scores_babyai}
\fontsize{8pt}{8.5pt}\selectfont
\begin{center}
\begin{sc}
\begin{tabular}{lcccc}
\toprule
Task & Random agent & Expert & JAT (raw) & JAT (normalized) \\ \midrule
Action Obj Door & 0.37 ± 0.39 & 0.99 ± 0.01 & 0.94 ± 0.14 & 0.93 ± 0.23 \\ 
Blocked Unlock Pickup & 0.00 ± 0.02 & 0.95 ± 0.01 & 0.95 ± 0.01 & 1.00 ± 0.01 \\ 
Boss Level & 0.06 ± 0.21 & 0.94 ± 0.05 & 0.52 ± 0.43 & 0.53 ± 0.49 \\ 
Boss Level No Unlock & 0.06 ± 0.19 & 0.94 ± 0.05 & 0.48 ± 0.43 & 0.48 ± 0.48 \\ 
Find Obj S5 & 0.08 ± 0.23 & 0.95 ± 0.04 & 0.95 ± 0.04 & 1.01 ± 0.05 \\ 
Go To & 0.13 ± 0.29 & 0.92 ± 0.07 & 0.83 ± 0.27 & 0.89 ± 0.34 \\ 
Go To Door & 0.45 ± 0.38 & 0.99 ± 0.00 & 0.99 ± 0.02 & 0.99 ± 0.04 \\ 
Go To Imp Unlock & 0.07 ± 0.22 & 0.83 ± 0.13 & 0.60 ± 0.41 & 0.70 ± 0.54 \\ 
Go To Local & 0.16 ± 0.30 & 0.93 ± 0.04 & 0.88 ± 0.14 & 0.94 ± 0.19 \\ 
Go To Obj & 0.13 ± 0.27 & 0.93 ± 0.03 & 0.93 ± 0.03 & 1.00 ± 0.04 \\ 
Go To Obj Door & 0.53 ± 0.39 & 0.99 ± 0.01 & 0.96 ± 0.10 & 0.94 ± 0.21 \\ 
Go To Red Ball & 0.17 ± 0.30 & 0.93 ± 0.04 & 0.92 ± 0.05 & 0.99 ± 0.07 \\ 
Go To Red Ball Grey & 0.12 ± 0.27 & 0.92 ± 0.05 & 0.91 ± 0.07 & 0.99 ± 0.08 \\ 
Go To Red Ball No Dists & 0.14 ± 0.28 & 0.93 ± 0.03 & 0.93 ± 0.03 & 1.00 ± 0.04 \\ 
Go To Red Blue Ball & 0.12 ± 0.27 & 0.92 ± 0.05 & 0.88 ± 0.11 & 0.95 ± 0.14 \\ 
Go To Seq & 0.08 ± 0.23 & 0.94 ± 0.05 & 0.72 ± 0.34 & 0.74 ± 0.40 \\ 
Key Corridor & 0.00 ± 0.00 & 0.91 ± 0.01 & 0.86 ± 0.16 & 0.94 ± 0.18 \\ 
Mini Boss Level & 0.07 ± 0.21 & 0.89 ± 0.10 & 0.61 ± 0.39 & 0.65 ± 0.47 \\ 
Move Two Across S8N9 & 0.00 ± 0.00 & 0.96 ± 0.01 & 0.02 ± 0.13 & 0.02 ± 0.13 \\ 
One Room S8 & 0.08 ± 0.21 & 0.92 ± 0.03 & 0.92 ± 0.04 & 1.00 ± 0.04 \\ 
Open & 0.10 ± 0.24 & 0.95 ± 0.05 & 0.94 ± 0.11 & 0.98 ± 0.13 \\ 
Open Door & 0.23 ± 0.34 & 0.99 ± 0.00 & 0.99 ± 0.00 & 1.00 ± 0.01 \\ 
Open Doors Order N4 & 0.16 ± 0.30 & 0.99 ± 0.01 & 0.95 ± 0.17 & 0.95 ± 0.20 \\ 
Open Red Door & 0.08 ± 0.21 & 0.92 ± 0.03 & 0.91 ± 0.03 & 1.00 ± 0.04 \\ 
Open Two Doors & 0.08 ± 0.20 & 0.98 ± 0.00 & 0.98 ± 0.00 & 1.00 ± 0.00 \\ 
Pickup & 0.08 ± 0.22 & 0.92 ± 0.07 & 0.76 ± 0.32 & 0.82 ± 0.38 \\ 
Pickup Above & 0.02 ± 0.09 & 0.91 ± 0.07 & 0.90 ± 0.08 & 0.99 ± 0.09 \\ 
Pickup Dist & 0.10 ± 0.24 & 0.86 ± 0.21 & 0.90 ± 0.07 & 1.05 ± 0.09 \\ 
Pickup Loc & 0.08 ± 0.23 & 0.91 ± 0.04 & 0.86 ± 0.16 & 0.94 ± 0.19 \\ 
Put Next S7N4 & 0.00 ± 0.03 & 0.96 ± 0.01 & 0.85 ± 0.22 & 0.88 ± 0.23 \\ 
Put Next Local & 0.00 ± 0.05 & 0.92 ± 0.03 & 0.63 ± 0.36 & 0.69 ± 0.40 \\ 
Synth & 0.11 ± 0.26 & 0.93 ± 0.06 & 0.77 ± 0.33 & 0.80 ± 0.40 \\ 
Synth Loc & 0.13 ± 0.29 & 0.94 ± 0.06 & 0.79 ± 0.33 & 0.81 ± 0.40 \\ 
Synth Seq & 0.07 ± 0.20 & 0.95 ± 0.04 & 0.52 ± 0.44 & 0.52 ± 0.50 \\ 
Unblock Pickup & 0.08 ± 0.22 & 0.91 ± 0.08 & 0.74 ± 0.32 & 0.79 ± 0.39 \\ 
Unlock & 0.03 ± 0.15 & 0.87 ± 0.10 & 0.52 ± 0.42 & 0.58 ± 0.50 \\ 
Unlock Local & 0.01 ± 0.09 & 0.98 ± 0.01 & 0.98 ± 0.01 & 1.00 ± 0.01 \\ 
Unlock Pickup & 0.00 ± 0.00 & 0.75 ± 0.04 & 0.76 ± 0.04 & 1.01 ± 0.05 \\ 
Unlock To Unlock & 0.00 ± 0.00 & 0.96 ± 0.00 & 0.80 ± 0.35 & 0.83 ± 0.37 \\ \bottomrule
\end{tabular}
\end{sc}
\end{center}
\end{table}

\clearpage

\begin{table}[p]
\caption{Comparison of performance scores across tasks on Meta-World. The table presents the episodic return (score) achieved by a random agent (averaged over 1,000 episodes), scores of the expert agent (as averaged from the dataset), scores of the learned agent, and the expert normalized score calculated as $\frac{{\text{{score}} - \text{{random\_score}}}}{{\text{{expert\_score}} - \text{{random\_score}}}}$.}
\label{tab:comprehensive_scores_metaworld}
\fontsize{8pt}{8.5pt}\selectfont
\begin{center}
\begin{sc}
\begin{tabular}{lcccc}
\toprule
Task & Random agent & Expert & JAT (raw) & JAT (normalized) \\ \midrule
Assembly & 45.3 ± 4.1 & 246.0 ± 3.5 & 238.0 ± 34.6 & 0.96 ± 0.17 \\ 
Basketball & 2.8 ± 1.2 & 628.0 ± 2.0 & 1.6 ± 0.4 & -0.00 ± 0.00 \\ 
BinPicking & 1.9 ± 0.4 & 425.6 ± 101.9 & 200.0 ± 222.1 & 0.47 ± 0.52 \\ 
Box Close & 76.4 ± 17.9 & 512.5 ± 107.8 & 462.6 ± 172.0 & 0.89 ± 0.39 \\ 
Button Press & 31.7 ± 5.2 & 643.1 ± 12.8 & 560.0 ± 182.6 & 0.86 ± 0.30 \\ 
Button Press Topdown & 29.0 ± 10.4 & 490.2 ± 27.2 & 266.2 ± 77.2 & 0.51 ± 0.17 \\ 
Button Press Topdown Wall & 29.0 ± 10.5 & 497.2 ± 31.4 & 275.8 ± 88.6 & 0.53 ± 0.19 \\ 
Button Press Wall & 9.0 ± 4.0 & 675.4 ± 15.0 & 638.3 ± 123.0 & 0.94 ± 0.18 \\ 
Coffee Button & 31.7 ± 6.4 & 731.1 ± 29.3 & 298.0 ± 285.6 & 0.38 ± 0.41 \\ 
Coffee Pull & 4.1 ± 0.4 & 259.9 ± 88.5 & 41.0 ± 69.8 & 0.14 ± 0.27 \\ 
Coffee Push & 4.2 ± 0.8 & 496.8 ± 118.2 & 153.1 ± 218.9 & 0.30 ± 0.44 \\ 
Dial Turn & 29.6 ± 16.7 & 793.6 ± 80.1 & 758.4 ± 120.4 & 0.95 ± 0.16 \\ 
Disassemble & 40.3 ± 7.5 & 42.8 ± 6.3 & 40.7 ± 9.9 & 0.17 ± 3.91 \\ 
Door Close & 5.3 ± 1.3 & 529.7 ± 27.2 & 524.3 ± 33.2 & 0.99 ± 0.06 \\ 
Door Lock & 112.3 ± 28.6 & 811.5 ± 34.1 & 696.3 ± 198.6 & 0.84 ± 0.28 \\ 
Door Open & 56.4 ± 11.2 & 581.9 ± 19.7 & 577.5 ± 53.7 & 0.99 ± 0.10 \\ 
Door Unlock & 94.2 ± 15.6 & 802.9 ± 17.1 & 768.3 ± 91.8 & 0.95 ± 0.13 \\ 
Drawer Close & 116.7 ± 253.1 & 867.9 ± 4.5 & 596.7 ± 223.4 & 0.64 ± 0.30 \\ 
Drawer Open & 126.8 ± 25.2 & 493.0 ± 2.5 & 485.9 ± 36.8 & 0.98 ± 0.10 \\ 
Faucet Close & 253.1 ± 22.9 & 753.9 ± 13.4 & 367.8 ± 91.1 & 0.23 ± 0.18 \\ 
Faucet Open & 244.1 ± 23.3 & 705.8 ± 7.1 & 566.1 ± 169.9 & 0.70 ± 0.37 \\ 
Hammer & 95.3 ± 9.0 & 693.2 ± 34.6 & 667.7 ± 89.4 & 0.96 ± 0.15 \\ 
Hand Insert & 2.8 ± 3.5 & 740.5 ± 36.7 & 688.1 ± 187.7 & 0.93 ± 0.25 \\ 
Handle Press & 80.4 ± 110.2 & 855.9 ± 72.7 & 735.0 ± 252.0 & 0.84 ± 0.32 \\ 
Handle Press Side & 57.0 ± 39.5 & 861.1 ± 20.0 & 64.5 ± 73.7 & 0.01 ± 0.09 \\ 
Handle Pull & 10.3 ± 13.5 & 669.4 ± 24.8 & 556.6 ± 161.7 & 0.83 ± 0.25 \\ 
Handle Pull Side & 2.1 ± 2.8 & 384.7 ± 102.9 & 195.1 ± 187.2 & 0.50 ± 0.49 \\ 
Lever Pull & 60.3 ± 15.8 & 612.0 ± 38.9 & 280.9 ± 234.8 & 0.40 ± 0.43 \\ 
Peg Insert Side & 1.7 ± 0.4 & 315.2 ± 140.1 & 254.3 ± 158.4 & 0.81 ± 0.51 \\ 
Peg Unplug Side & 4.7 ± 2.8 & 456.1 ± 81.7 & 80.6 ± 145.5 & 0.17 ± 0.32 \\ 
Pick Out Of Hole & 1.5 ± 0.2 & 219.6 ± 88.9 & 2.1 ± 0.1 & 0.00 ± 0.00 \\ 
Pick Place & 1.6 ± 1.0 & 419.1 ± 98.2 & 135.8 ± 200.1 & 0.32 ± 0.48 \\ 
Pick Place Wall & 0.0 ± 0.0 & 450.6 ± 64.1 & 43.7 ± 129.7 & 0.10 ± 0.29 \\ 
Plate Slide & 74.6 ± 13.8 & 527.0 ± 155.3 & 481.5 ± 190.2 & 0.90 ± 0.42 \\ 
Plate Slide Back & 33.5 ± 11.2 & 718.2 ± 87.4 & 196.9 ± 1.7 & 0.24 ± 0.00 \\ 
Plate Slide Back Side & 34.3 ± 11.5 & 729.6 ± 69.1 & 703.7 ± 117.3 & 0.96 ± 0.17 \\ 
Plate Slide Side & 22.6 ± 17.4 & 662.8 ± 102.8 & 122.6 ± 24.6 & 0.16 ± 0.04 \\ 
Push & 5.5 ± 2.4 & 750.6 ± 44.0 & 702.4 ± 157.6 & 0.94 ± 0.21 \\ 
Push Back & 1.2 ± 0.2 & 85.0 ± 107.1 & 82.2 ± 108.0 & 0.97 ± 1.29 \\ 
Push Wall & 6.1 ± 3.2 & 748.9 ± 10.6 & 158.8 ± 224.6 & 0.21 ± 0.30 \\ 
Reach & 149.7 ± 44.7 & 681.4 ± 133.7 & 332.2 ± 171.5 & 0.34 ± 0.32 \\ 
Reach Wall & 143.3 ± 36.6 & 746.1 ± 104.2 & 631.6 ± 224.0 & 0.81 ± 0.37 \\ 
Shelf Place & 0.0 ± 0.0 & 241.3 ± 24.6 & 92.1 ± 112.0 & 0.38 ± 0.46 \\ 
Soccer & 5.7 ± 4.6 & 375.2 ± 140.2 & 291.6 ± 161.8 & 0.77 ± 0.44 \\ 
Stick Pull & 2.6 ± 1.4 & 523.6 ± 18.9 & 480.1 ± 119.3 & 0.92 ± 0.23 \\ 
Stick Push & 2.8 ± 1.0 & 627.9 ± 10.2 & 303.2 ± 298.8 & 0.48 ± 0.48 \\ 
Sweep & 11.2 ± 7.3 & 494.8 ± 43.3 & 16.7 ± 18.7 & 0.01 ± 0.04 \\ 
Sweep Into & 12.5 ± 10.7 & 799.2 ± 19.1 & 793.3 ± 47.5 & 0.99 ± 0.06 \\ 
Window Close & 57.5 ± 7.1 & 591.3 ± 38.6 & 414.3 ± 207.4 & 0.67 ± 0.39 \\ 
Window Open & 43.4 ± 2.1 & 590.8 ± 57.1 & 577.3 ± 63.9 & 0.98 ± 0.12 \\ \bottomrule 
\end{tabular}
\end{sc}
\end{center}
\end{table}

\clearpage

\begin{table}[p]
\caption{Comparison of performance scores across tasks on MuJoCo. The table presents the episodic return (score) achieved by a random agent (averaged over 1,000 episodes), scores of the expert agent (as averaged from the dataset), scores of the learned agent, and the expert normalized score calculated as $\frac{{\text{{score}} - \text{{random\_score}}}}{{\text{{expert\_score}} - \text{{random\_score}}}}$.}
\label{tab:comprehensive_scores_mujoco}
\fontsize{8pt}{8.5pt}\selectfont
\begin{center}
\begin{sc}
\begin{tabular}{lcccc}
\toprule
Task & Random agent & Expert & JAT (raw) & JAT (normalized) \\ \midrule
Ant & -59.9 ± 99.6 & 5846.4 ± 942.6 & 5110.5 ± 1720.8 & 0.88 ± 0.29 \\ 
Inverted Double Pendulum & 57.5 ± 17.5 & 9338.7 ± 352.6 & 8663.7 ± 1259.4 & 0.93 ± 0.14 \\ 
Half Cheetah & -285.0 ± 79.8 & 7437.8 ± 173.3 & 6595.9 ± 244.4 & 0.89 ± 0.03 \\ 
Hopper & 18.4 ± 17.1 & 1858.7 ± 534.1 & 1409.0 ± 385.6 & 0.76 ± 0.21 \\ 
Humanoid & 122.0 ± 35.3 & 6281.0 ± 1795.8 & 712.6 ± 120.6 & 0.10 ± 0.02 \\ 
Inverted Pendulum & 6.1 ± 3.5 & 475.4 ± 179.0 & 117.4 ± 22.0 & 0.24 ± 0.05 \\ 
Pusher & -149.7 ± 7.4 & -25.2 ± 6.7 & -25.0 ± 6.3 & 1.00 ± 0.05 \\ 
Reacher & -43.0 ± 3.9 & -5.7 ± 2.5 & -5.9 ± 2.4 & 0.99 ± 0.06 \\ 
Humanoid Standup & 33135.8 ± 2481.9 & 273574.2 ± 85253.3 & 116736.7 ± 22765.5 & 0.35 ± 0.09 \\ 
Swimmer & 0.8 ± 10.7 & 92.2 ± 4.4 & 94.0 ± 4.1 & 1.02 ± 0.04 \\ 
Walker 2d & 2.7 ± 6.1 & 4631.2 ± 1059.0 & 4381.3 ± 851.1 & 0.95 ± 0.18 \\ \bottomrule
\end{tabular}
\end{sc}
\end{center}
\end{table}

\clearpage
\section{JAT Dataset In Depth}
\label{appendix:dataset}

\subsection{Sequential Decision-Making Datasets}

For each decision-making environment, we collect a set of interactions using expert agents.
Detailed scores are available in the Appendix \ref{appendix:full_results}.

\paragraph{Atari}
We use the 57 games from the Arcade Learning Environment \citep[ALE]{bellemare2013arcade} as a benchmark in our research, amassing roughly 500,000 interactions per game. Episode lengths varied significantly depending on the specific game. For each game, we trained a dedicated agent for 2 billion steps using the asynchronous implementation of Proximal Policy Optimization \citep{schulman2017proximal} from Sample Factory \citep{petrenko2020sample}. The expert agents achieve above human performance on 43 tasks\footnote{The expert score is below the human score for Asterix, Bowling, Centipede, Fishing Derby, Kangaroo, Montezuma's Revenge, Ms. Pacman, Pitfall, Private Eye, River Raid, Seaquest, Skiing, Solaris, Venture}. 

\paragraph{BabyAI}
BabyAI stands out in our study due to its unique characteristic of being partially observable and its dual-modality observations \citep{chevalier_boisvert2019babyai, chevalier_boisvert2023minigrid}.
Using the bot provided with the BabyAI paper \citep{chevalier_boisvert2019babyai}, we gathered 100,000 episodes for 39 of its available settings. Each interaction consists of a text observation (mission), a discrete observation ($7\times 7$ symbolic representation of the agent's field of view), an action, and a reward.

\paragraph{Meta-World}
Meta-World's MT50 benchmark provides a set of 50 diverse and challenging robot manipulation tasks \citep{yu2019meta}. Similar to the methodology used for Atari, we trained one agent per task using the asynchronous PPO \citep{schulman2017proximal} implementation of \citep{petrenko2020sample}. The trained agents solved most of the tasks, except for Assembly and Disassemble, where they failed to reach the expected performance. We limit the number of timesteps per episode to 100, which proved to be sufficient for solving the tasks. Without this limit, much of the subsequent dataset would consist of the stabilization phases of the agents after goal attainment, reducing its relevance. We then used the trained agents to generate 10,000 episodes per environment.

\paragraph{MuJoCo}
We included the MuJoCo locomotion benchmark suite \citep{todorov2012mujoco, brockman2016openai} comprising 11 continuous control tasks into our study due to its diverse challenges in domain complexity and task difficulty, and its wide recognition in the research literature. Following our methodologies for Atari and Meta-World, we individually trained agents for each task using asynchronous PPO \citep{schulman2017proximal} from Sample Factory \citep{petrenko2020sample}. These agents successfully solved all tasks, achieving scores that meet or exceed the current highest standards. Subsequently, we employed these agents to generate 10,000 episodes per environment.

\begin{figure}[h]
    \centering
    \includegraphics[width=\textwidth]{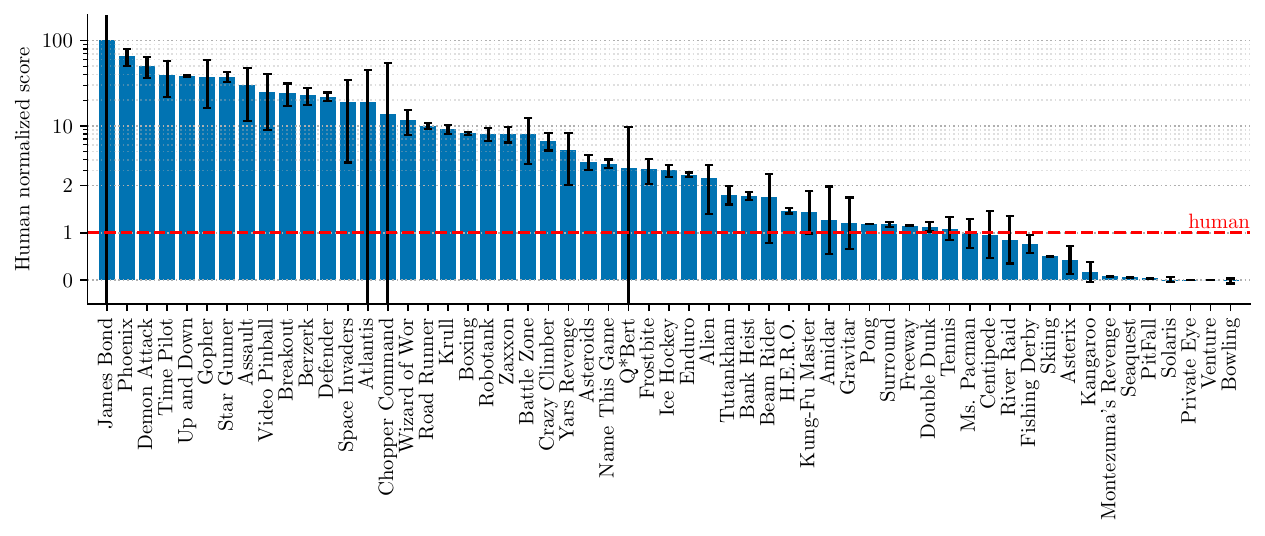}
    \caption{Human normalized dataset scores for Atari.}
    \label{fig:human_normalized_atari_dataset}
\end{figure}

Each sample in this dataset is an episode. This episode consists of a list of observations, actions and rewards, the nature and size of which depend on the task. In Figure \ref{fig:human_normalized_atari_dataset}, we represent for each Atari game the average return of the episodes of the dataset normalized by the human score from \citep{mnih2015human}.
Notably, for 43 games the average score is higher than the human score, and for 31 games the average score is more than twice the human score. It should be noted, however, that for 7 games (Bowling, Montezuma's Revenge, PitFall, Private Eye, Seaquest, Solaris and Venture) the average score is less than 10\% of the human score.

\begin{figure}
    \centering
    \includegraphics[width=0.992\textwidth]{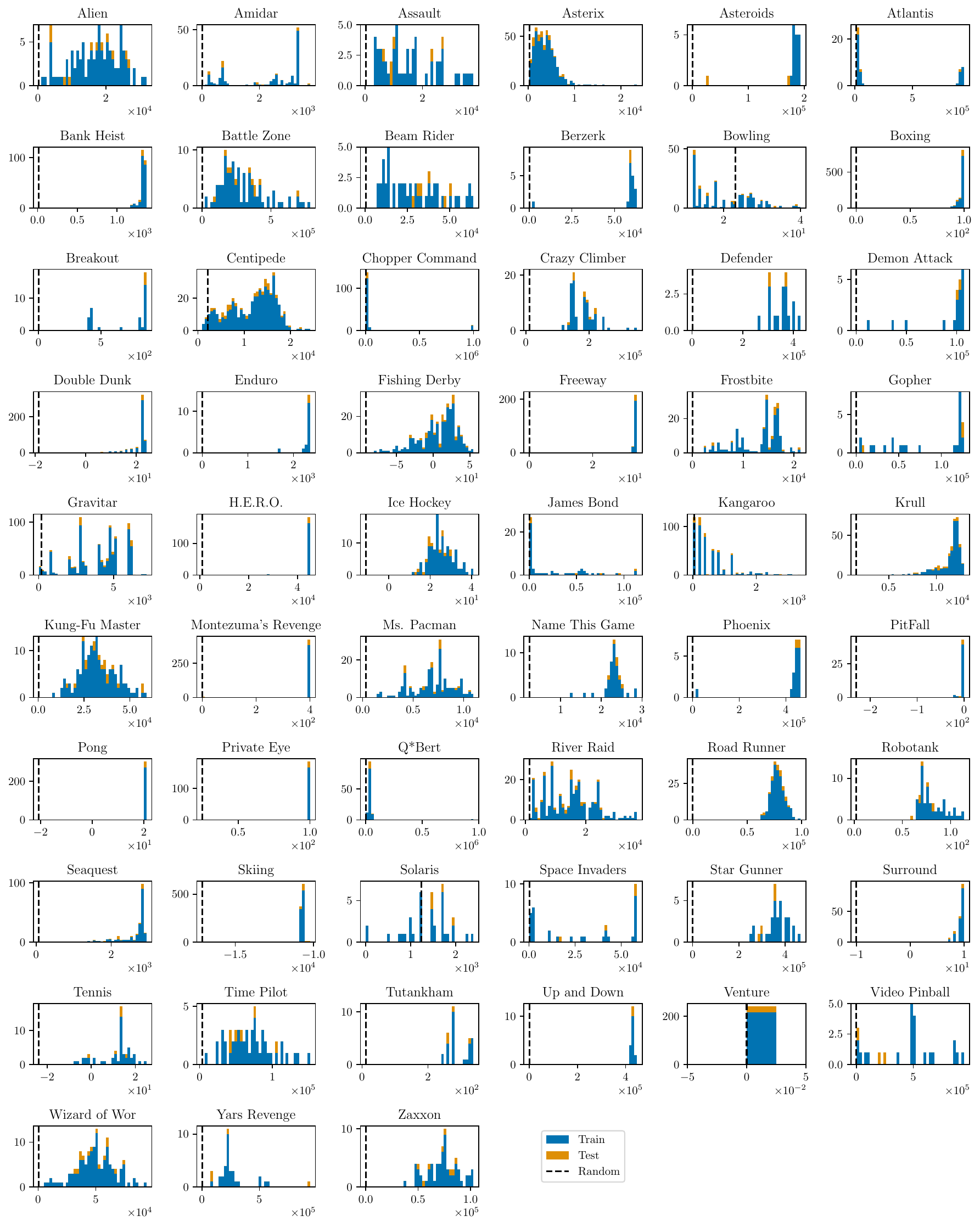}
    \caption{Atari dataset return distribution.}
    \label{fig:dataset_atari_return_distribution}
\end{figure}

\begin{figure}
    \centering
    \includegraphics[width=\textwidth]{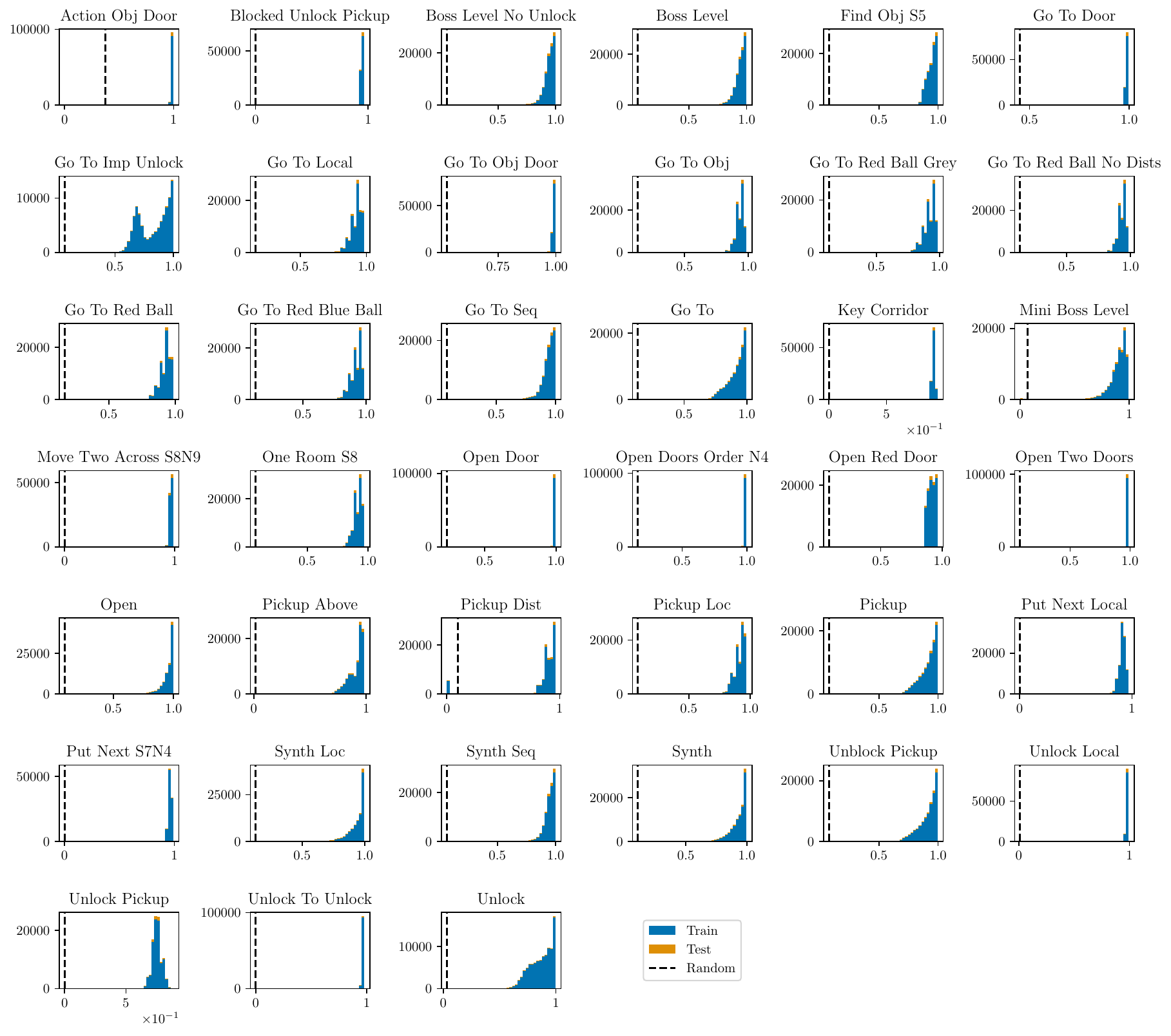}
    \caption{BabyAI dataset return distribution.}
    \label{fig:dataset_babyai_return_distribution}
\end{figure}

\begin{figure}
    \centering
    \includegraphics[width=\textwidth]{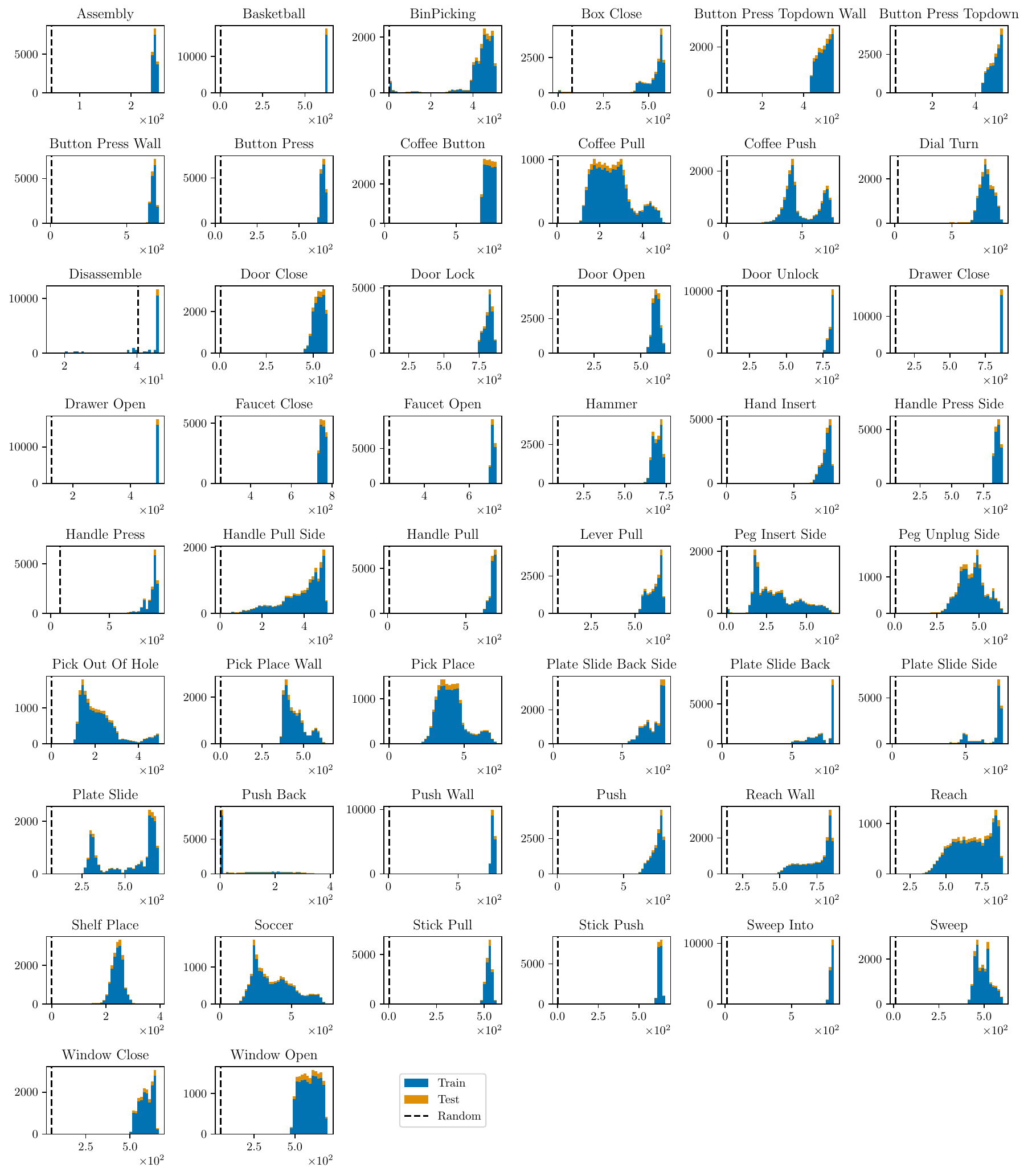}
    \caption{Meta-World dataset return distribution.}
    \label{fig:dataset_metaworld_return_distribution}
\end{figure}

\begin{figure}
    \centering
    \includegraphics[width=\textwidth]{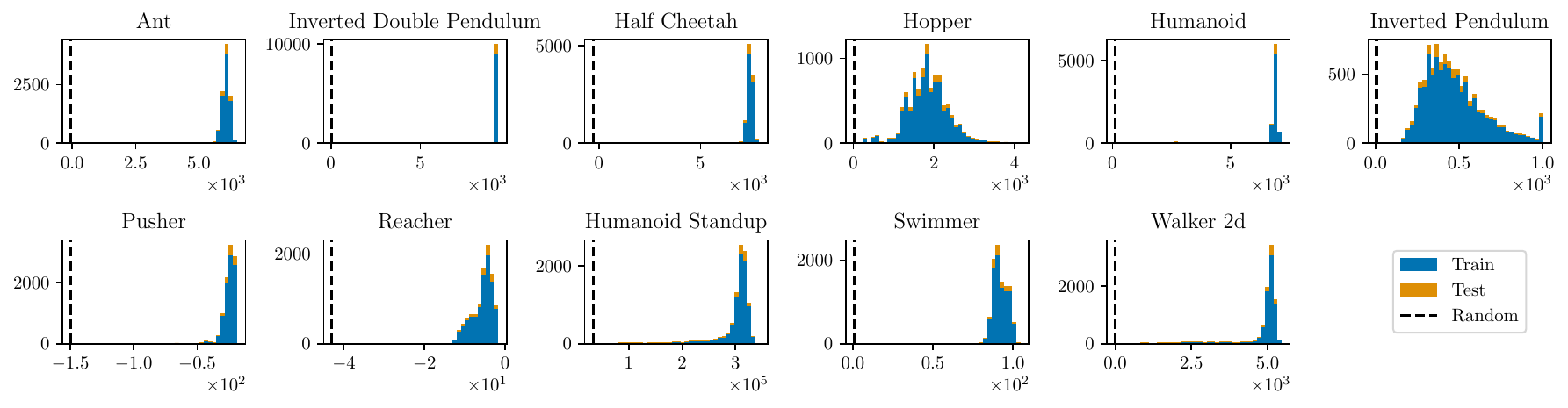}
    \caption{MuJoCo dataset return distribution.}
    \label{fig:dataset_mujoco_return_distribution}
\end{figure}

We also plot the distribution of returns for each task, which provides a more detailed picture than a simple average. Figure \ref{fig:dataset_atari_return_distribution} shows this distribution for Atari, Figure \ref{fig:dataset_babyai_return_distribution} for BabyAI, Figure \ref{fig:dataset_metaworld_return_distribution} for Meta-World and Figure \ref{fig:dataset_mujoco_return_distribution} for MuJoCo.

\clearpage
\subsubsection{Text-Centric Datasets}

\paragraph{Oscar}
Common Crawl-based text documents have been widely used in the past to create datasets for Language Modeling \citep{radford2019language, brown2020language, raffel2020exploring}. We chose to leverage the unshuffled deduplicated English subset of the OSCAR\footnote{Its original version from 2019: \url{https://huggingface.co/datasets/oscar}} corpus \citep{ortiz_su_arez2020monolingual} for our language modeling objective. As such crawled internet data needs to be cleaned before using it for training Language Models (e.g. deduplication, filtering out machine-generating content), we reused both the cleaning and deduplication pipeline from the ROOTS corpus \citep{laurençon2023bigscience}. The initial dataset was shuffled, split into a training (95\%) and test (5\%) set, and evenly split into 30 shards on which the cleaning and deduplication pipelines were applied to reduce the memory needs. Shards were then concatenated back together, leading to a final dataset of 245 million documents (compared to 304 million documents in the initial dataset). 

\paragraph{Conceptual-Captions}
We include the Conceptual-Captions dataset \citep{sharma2018conceptual}, as it is a key resource for image captioning and visual understanding tasks. It contains over 2.6 million training examples and over 12,000 test examples, with a wide range of web-sourced images, each paired with a descriptive caption. 

\paragraph{OK-VQA}
We include the OK-VQA dataset \citep{marino2019ok} because it is an essential resource for visual question answering tasks that focus on the intersection of visual perception and knowledge-based reasoning. With over 14,000 samples, it contains a wide range of images, each associated with questions that require not only visual understanding, but also external knowledge for an accurate answer.

\paragraph{Wikipedia}
The Wikipedia dataset, built from the Wikipedia dump \citep{wikidump}, contains over 6 million English language samples as of March 1, 2022. It offers a wide range of topics and a wealth of information. By using this dataset, we aim to improve the language processing capabilities of our model and provide access to extensive reservoir of encyclopedic knowledge.

\clearpage
\section{Image captioning additional examples}
\label{appendix:image_captioning}
\begin{figure*}[ht]
\centering
\begin{tabular}{C{5.1cm}C{5.1cm}C{5.1cm}}
\includegraphics[width=0.17\textwidth]{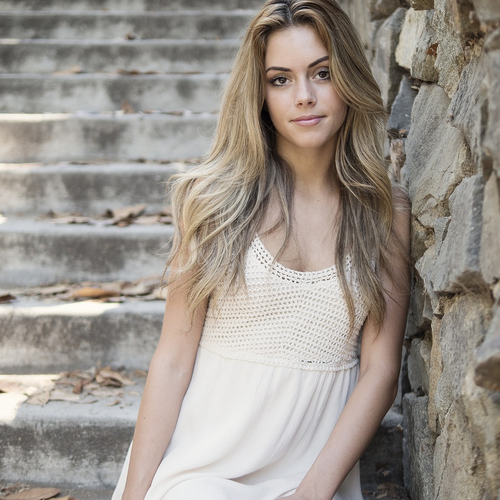} & \includegraphics[width=0.17\textwidth]{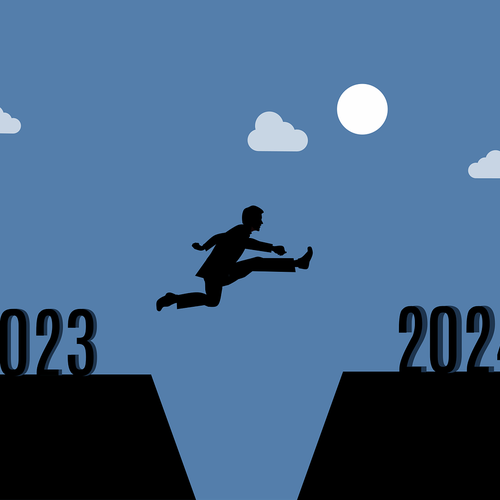} & \includegraphics[width=0.17\textwidth]{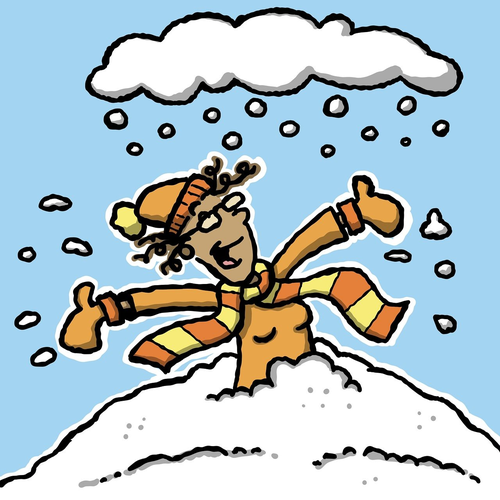}\\
\textcolor{completioncolor}{tattoo on the left inner forearm. $\sim$ artist. $\sim$ $\sim$ photo sharing website} &
\textcolor{completioncolor}{ - shaped cloud formation over a city. $\sim$ photo by person. \#zn. - \#zn.0 \# christmas } &
\textcolor{completioncolor}{ and illustration of the new year. photo by person.} \\ \addlinespace
\includegraphics[width=0.17\textwidth]{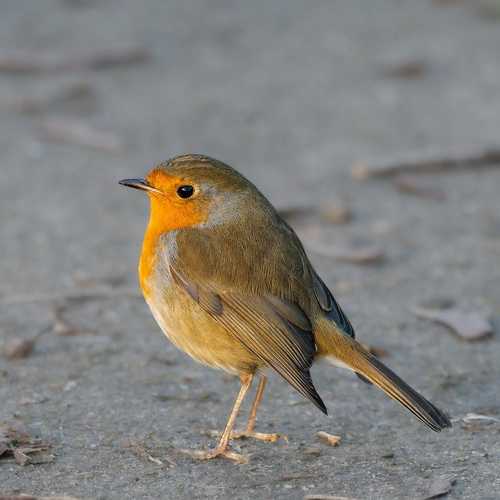} & \includegraphics[width=0.17\textwidth]{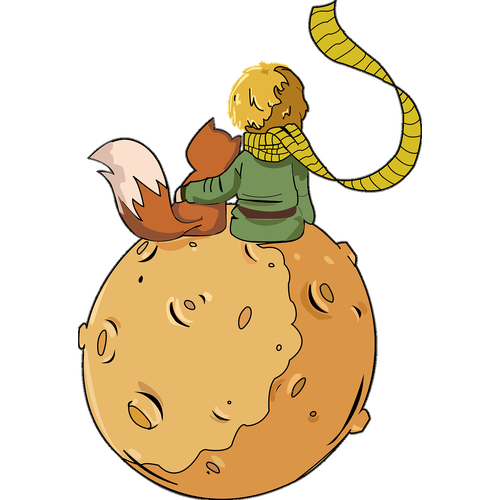} & \includegraphics[width=0.17\textwidth]{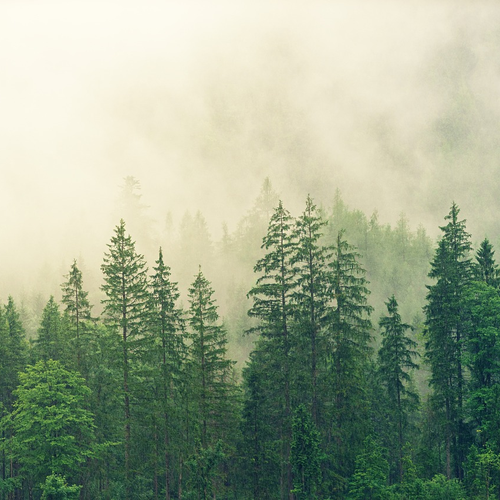}\\
\textcolor{completioncolor}{s and drawings on the ceiling of a building.} &
\textcolor{completioncolor}{ - cut emerald - cut diamonds are a perfect addition to any home.} &
\textcolor{completioncolor}{ day in the green forest.} \\ \addlinespace
\includegraphics[width=0.17\textwidth]{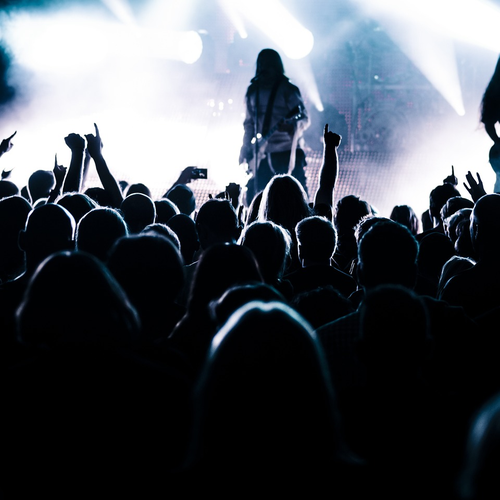} & \includegraphics[width=0.17\textwidth]{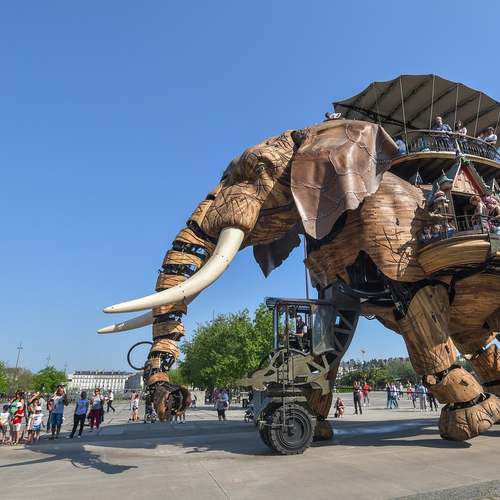} & \includegraphics[width=0.17\textwidth]{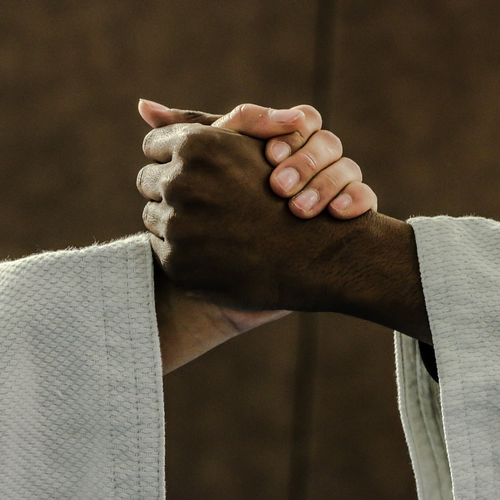}\\
\textcolor{completioncolor}{ for the first time! by person, the man who is now on the right.} &
\textcolor{completioncolor}{s are part of the annual event. organisation} &
\textcolor{completioncolor}{ging it up : the model was spotted wearing a pair of black jeans, a white t - shirt and black t - shirt} \\ \addlinespace
\includegraphics[width=0.17\textwidth]{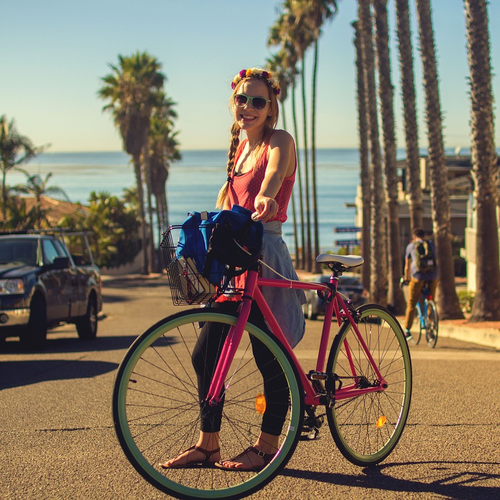} & \includegraphics[width=0.17\textwidth]{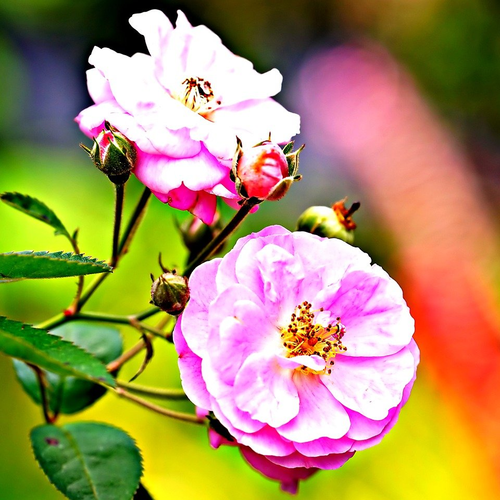}& \includegraphics[width=0.17\textwidth]{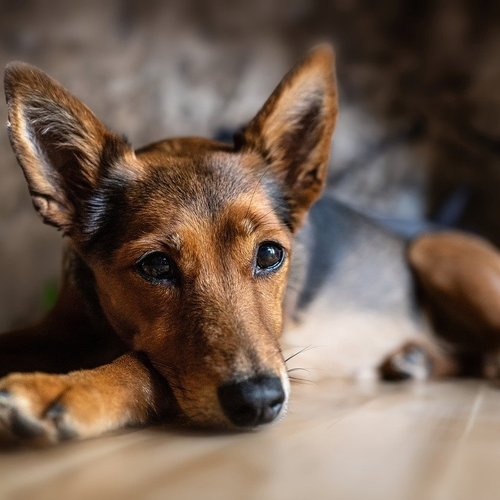}\\
\textcolor{completioncolor}{s and other souvenirs at the market.} &
\textcolor{completioncolor}{s of the day : flowers} &
\textcolor{completioncolor}{dog in a bedroom.} \\
\end{tabular}
\caption{JAT image captioning additional examples.}
\label{fig:image_captioning_2}
\end{figure*}

\clearpage
\section{Reward as a Task Determinant}
\label{appendix:reward_task_determinant}

In multi-task learning, different environments may share identical dynamics and observational structures while differing in their ultimate goals (i.e., reward functions). Initially, the agent cannot distinguish the specific task it is facing.
In most cases, this problem does not arise. For BabyAI, for example, the goal is an explicit part of the observation. For Atari, a single frame is sufficient to determine the game, and therefore the goal. In our dataset, the only domain that could be challenging in this respect is Meta-World, for which the structure of observations and dynamics is consistent across tasks. Note also that even in this case, it should be possible for the agent in some instances to infer the task from the initial conditions. We confirm this hypothesis in the following experiment.

To solve the problem of task indeterminacy, Gato introduces a method of pre-empting the sequence with an expert demonstration (prompt) to guide the agent. While this approach is effective, it imposes an important limitation: a demonstration must be available, and this demonstration must be sufficiently complete to clearly define the task. In the JAT model, we adopt a less restrictive and simpler approach by incorporating the reward signal directly into the observation encoding. We believe that this integration can, in most cases, provide the agent with sufficient context to remove ambiguity about the task at hand.

\begin{figure}[ht]
\vskip 0.2in
\begin{center}
\begin{subfigure}[t]{0.5\textwidth}
\centerline{\includegraphics[width=\textwidth]{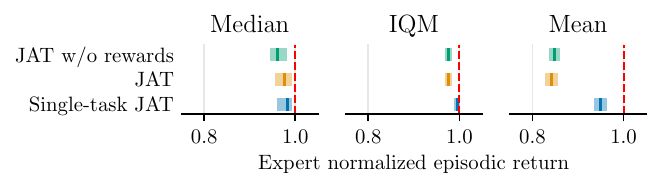}}
\caption{Aggregated metrics.}
\label{subfig:agregated_reward_ablation}
\end{subfigure}
\begin{subfigure}[t]{0.4\textwidth}
\centerline{\includegraphics[width=\textwidth]{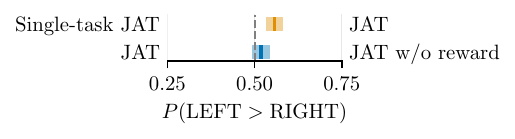}}
\caption{Probability of improvement.}
\label{subfig:prob_of_improvement_reward_ablation}
\end{subfigure}
\caption{Results of the reward ablation. The vertical bars are the estimated values and the shaded areas are the 95\% stratified bootstrap CIs. The experiments were conducted on a selection of 10 tasks from the Meta-World benchmark. Displayed are the results of an ablation study on our JAT model variations: Single-task JAT with each task learned by a dedicated agent; JAT without rewards where the training omits reward signals; and the full JAT model integrating reward signals. Results are based on 100 evaluations per task.}
\label{fig:reward_ablation}
\end{center}
\vskip -0.2in
\end{figure}

To support our hypothesis on the effectiveness of integrating reward signals into observations, we conducted an experiment with three different settings. First, to create a baseline where task indeterminacy is absent, we trained individual agents, each on a specific task from a random subset of 10 Meta-World tasks. This single-task training ensures that each agent is perfectly matched to its respective task, without any ambiguity. Next, we introduced a degree of indeterminacy by training a single model on the same 10 tasks without access to the reward signal, presenting a scenario that simulates a worst-case uncertainty condition. We compare these two settings with our full JAT model, i.e. with access to the reward signal, trained on the same selection of tasks. We compared the performance of these three scenarios, with the results detailed in Figures \ref{fig:reward_ablation} following the recommendations of \citep{agarwal2021deep}.


Firstly, it's notable that the JAT model trained on a single task surpasses other settings, thus demonstrating the existence of a negative impact of task indeterminacy. However, this impact is actually very minor, and even in the most unfavorable setting (JAT without reward), the normalized IQM score reaches 97.6 ± 0.7\%. This confirms the previously formulated intuition that the task can generally be inferred from the initial conditions.
Then, when comparing the JAT model with and without access to the reward, we observe a probability of improvement from the former over the latter of 51.8 ± 2.5\%, indicating that the addition of the reward has a significant, albeit small, positive effect on resolving indeterminacy. Lastly, the most significant gap is observed in the average score. This can be attributed to the fact that this metric accounts for outliers. Here, the outliers are the tasks suffering from indeterminacy, for which the agent often fails to resolve the task.

In summary, the key insight from this study is that complex solutions like prompting are often not required to address this task indetermination issue, as it typically presents a minimal challenge. Furthermore, in instances where the problem does manifest, implementing a straightforward strategy like incorporating the reward into the observation proves to be an effective measure for mitigation.

\end{document}